\documentclass[10pt,twocolumn,letterpaper]{article}

\usepackage{cvpr}              
\usepackage{multirow}
\usepackage{bm}
\usepackage{natbib}
\setcitestyle{numbers,square}
\usepackage[accsupp]{axessibility}

\usepackage[dvipsnames]{xcolor}

\definecolor{cvprblue}{rgb}{0.21,0.49,0.74}
\definecolor{yxy}{rgb}{0.21,0.49,0.74}
\definecolor{pz}{rgb}{0.61,0.5,0.5}
\definecolor{shq}{rgb}{0.61,0.9,0.5}
\usepackage[pagebackref,breaklinks,colorlinks,citecolor=cvprblue]{hyperref}

\def\paperID{5935} 
\def\confName{CVPR}
\def\confYear{2024}

\title{3D Multi-frame Fusion for Video Stabilization}

\author{Zhan Peng\hspace{0.1in} 
        Xinyi Ye\hspace{0.1in} 
        Weiyue Zhao\hspace{0.1in} 
        Tianqi Liu\hspace{0.1in} 
        Huiqiang Sun\hspace{0.1in}
        Baopu Li\hspace{0.1in}
        Zhiguo Cao\footnotemark[1]~\hspace{0.1in} \\
 School of AIA, Huazhong University of Science and Technology\hspace{0.3in}\\
{\tt\small \{peng\_zhan,xinyiye,zhaoweiyue,tq\_liu,shq1031,zgcao\}@hust.edu.cn}\\
{\tt\small bpli.cuhk@gmail.com}\\
\vspace{-4mm}
}
\begin{document}

\maketitle

\renewcommand{\thefootnote}{\fnsymbol{footnote}} 
\footnotetext[1]{Corresponding author.}

\begin{abstract}
In this paper, we present RStab, a novel framework for video stabilization that integrates 3D multi-frame fusion through volume rendering.
Departing from conventional methods, we introduce a 3D multi-frame perspective to generate stabilized images, addressing the challenge of full-frame generation while preserving structure.
%
The core of our RStab framework lies in \textbf{S}tabilized \textbf{R}endering~(SR), a volume rendering module, fusing multi-frame information in 3D space.
Specifically, SR involves warping features and colors from multiple frames by projection, fusing them into descriptors to render the stabilized image.
However, the precision of warped information depends on the projection accuracy, a factor significantly influenced by dynamic regions. In response, we introduce the \textbf{A}daptive \textbf{R}ay \textbf{R}ange~(ARR) module to integrate depth priors, adaptively defining the sampling range for the projection process. Additionally, we propose \textbf{C}olor \textbf{C}orrection~(CC) assisting geometric constraints with optical flow for accurate color aggregation.
Thanks to the three modules, our RStab demonstrates superior performance compared with previous stabilizers in the field of view (FOV), image quality, and video stability across various datasets.
\end{abstract}

\section{Introduction}
With the widespread adoption of smartphones, videos have become an important medium for documenting and sharing lives. The videos captured with handheld devices often suffer from annoying shakes. To mitigate this prevalent issue, numerous researchers devote efforts to developing video stabilization algorithms. These methods typically involve three steps: camera trajectory estimation, trajectory smoothing, and stabilized frame generation.

\begin{figure}[t]
   \includegraphics[width=1\linewidth]{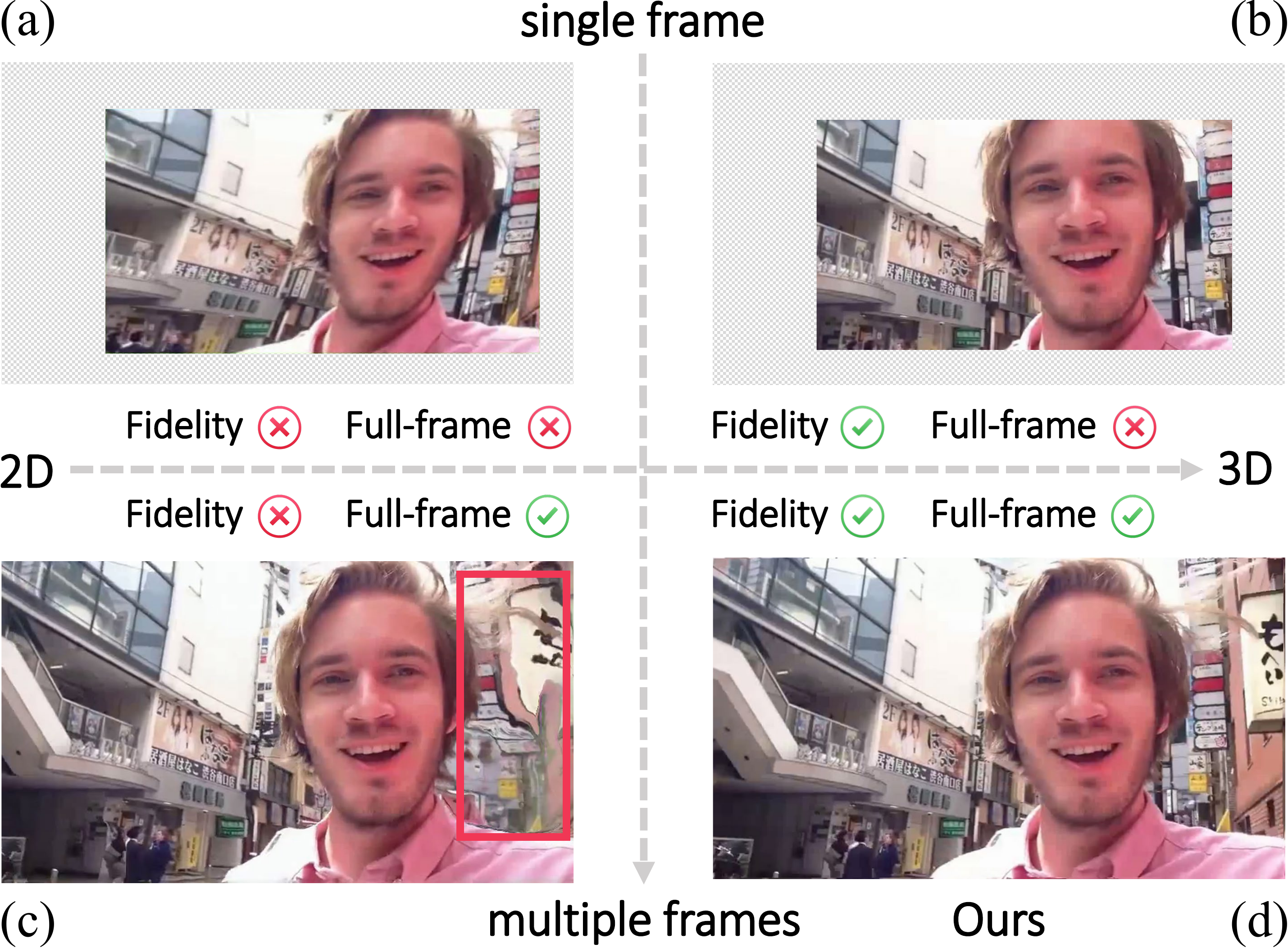}
\caption{
\textbf{Existing dilemmas and our method.} (a) and (b) exhibit cropping issues, characteristic of single-frame methods. (a) and (c) encounter difficulties in preserving structure, inherent in 2D-based approaches. Fortunately, our proposed method (d) not only mitigates distortion and artifacts but also maintains no-cropping stabilized frames.}
   
   \label{fig:fig1}
\end{figure}

To obtain a smooth image sequence, known as stabilized frames, early methods employ $2$D-plane transformations (homography~\cite{liu2013bundle,liu2017codingflow}, feature trajectories~\cite{lee2009feature,liu2014steadyflow,Yeong2015}, motion vectors~\cite{liu2011subspace}) on single frames. However, these methods suffer from two major problems. First, these single-frame approaches may produce notable missing regions at the boundary of generated stabilized images, requiring aggressive cropping to ensure a rectangular frame for video~(cropping in Fig.~\ref{fig:fig1}(a)), further resulting in a substantial reduction in the field of view (FOV). Second, $2$D transformations could give rise to structure distortion due to the lack of $3$D physical information~(shear in Fig.~\ref{fig:fig1}(a)).

In pursuit of the stabilized full-frame, recent 2D methods~\cite{choi2020difrint,liu2021fusta,zhao2023fast} leverage nearby frames to fill in the unseen content within the target frame. However, due to the inherent absence of physical constraints in $2$D transformations, 2D-based multiple-frame methods fail to preserve the structure, especially the parallax regions (Fig.~\ref{fig:fig1}(c)). 
To obtain the structure-preserved stabilized frame, some methods~\cite{liu2009content,lee2021deep3d,smith2009light,liu2012depth} leverage $3$D transformations to simulate real-world settings, employing camera poses and epipolar constraints to ensure the image structure. 
However, due to limited information from a single frame, they cannot generate a full frame, as shown in Fig.~\ref{fig:fig1}(b).
In brief, the ongoing challenge of concurrently addressing full-frame generation while preserving structure for video stabilization remains a major concern for most current research works.

To overcome the above problems, intuitively, employing multi-frame fusion with $3$D transformations could offer a promising solution. However, two issues may still hinder $3$D transformations from incorporating information from neighboring frames.
First, since view changes induce geometric deformation, the incorporated information from nearby frames may be inconsistent, suggesting that image blending, e.g., averaging, may lead to distortion. 
Second, videos feature dynamic objects across frames, which cannot be adequately modeled by $3$D constraints. The direct aggregation of information from nearby frames with $3$D projection results in a noticeable blur~(refer to the experiments).

Motivated by the above insights and analysis, we propose a video stabilization framework termed RStab for integrating multi-frame fusion and $3$D constraints to achieve full-frame generation and structure preservation.  
Specifically,  we propose \textbf{S}tabilized \textbf{R}endering~(SR), a $3$D multi-frame fusion module using volume rendering. 
Instead of simple image blending, SR employs both color and feature space to fuse nearby information into spatial descriptors for the scene geometry, such as volume densities of spatial points. 
Visible points usually come with high volume densities, exhibiting consistent textures in their projections across frames.
The observation suggests that points with higher consistency in aggregating information exhibit higher volume densities, implying a greater contribution to the final rendered color.

To mitigate the impacts of dynamic regions, we propose \textbf{A}daptive \textbf{R}ay \textbf{R}ange~(ARR) and \textbf{C}olor \textbf{C}orrection(CC) modules.
The introduction of multi-frame depth priors in ARR constrains the sampling range for spatial points around the surface of objects. 
A narrow sampling range around the surface decreases the risk of projecting spatial points onto dynamic regions, thereby suppressing the inconsistent information aggregation induced by the dynamic objects.
Despite ARR, colors are sensitive to projection inaccuracy, indicating a narrow range is insufficient.
Hence, we design CC to refine the projection for color aggregation. The core of CC lies in assisting geometry constraints with optical flow, which matches pixels with similar textures containing the color information.

By applying the three modules, RStab demonstrates the ability of full-frame generation with structure preservation~(Fig.~\ref{fig:fig1}(d)) and outperforms all previous video stabilization algorithms in FOV, image quality, and video stability across various datasets.
In summary, our key contributions are as follows:

\begin{itemize}[leftmargin=0.6cm]
\item We present a novel $3$D multi-frame fusion framework for video stabilization to render full-frame stabilized images with structure preservation. 
\item We propose Stabilized Rendering, which fuses multiple frames in both color and feature space. We augment Stabilized Rendering with the introduction of the Adaptive Ray Range module and Color Correction module, enhancing its capacity to address dynamic regions.
\item Our video stabilization framework, RStab, demonstrates state-of-the-art (SOTA) performance across various datasets.
\end{itemize}

\section{Related Work}
\noindent{\textbf{2D-based Video Stabilization.}} 
2D video stabilization algorithms model camera trajectory and generate stabilized frames through transformations on a 2D plane, including homography~\cite{grundmann2011l1,goldstein2012epipolar,liu2013bundle,yu2020stable,liu2017codingflow}, feature trajectories~\cite{lee2009feature,liu2014steadyflow,Yeong2015,yu2018selfie}, motion vectors~\cite{liu2011subspace,liu2016meshflow,jiang2017deformation,xu2022dut}, and dense flow~fields~\cite{yu2020stable,liu2021fusta,choi2020difrint,Zhao2020PWStableNet,chen2021pixstab}. Early methods~\cite{grundmann2011l1,goldstein2012epipolar} estimate global transformations, which proved inadequate for handling complex camera effects such as the parallax effect. Certain approaches estimate multiple local motions~\cite{liu2013bundle,liu2016meshflow} or pixel-wise warping field~\cite{yu2020stable, Zhao2020PWStableNet,xu2022dut} for a single image, offering some relief for the challenges encountered by global transformation methods. 
However, due to the limited information from a single frame, these methods may result in missing content in the stabilized video. To address this, some methods~\cite{liu2021fusta,zhao2023fast,choi2020difrint} fuse information from multiple neighboring frames, enabling full-frame generation. Despite achieving a full frame, the 2D transformations lack real-world physical constraints, leading to challenges in preserving image structure.

\noindent{\textbf{$3$D-based Video Stabilization.}}
$3$D-based video stabilizers model $3$D camera trajectory and stabilize frames with epipolar projection. Some methods~\cite{liu2009content,lee2021deep3d} rely on the video itself, warping images instructed by projection while preserving content. 
Others integrate specialized hardware, such as depth cameras~\cite{liu2012depth}, light field cameras~\cite{smith2009light}, gyroscopes~\cite{shi2022fused}, and IMU sensors~\cite{li2023imu}, to assist with scene geometry. 
Both kinds of stabilizers estimate the physical motion of the real world and introduce $3$D constraints in warping, benefiting stability and structure preservation.
However, relying on a single frame, $3$D-based video stabilizers have a limited field of view. To mitigate the issue, in this paper, we extend single-frame to multi-frame in $3$D space for video stabilization.

 \begin{figure*}[t]
   \includegraphics[width=1\linewidth]{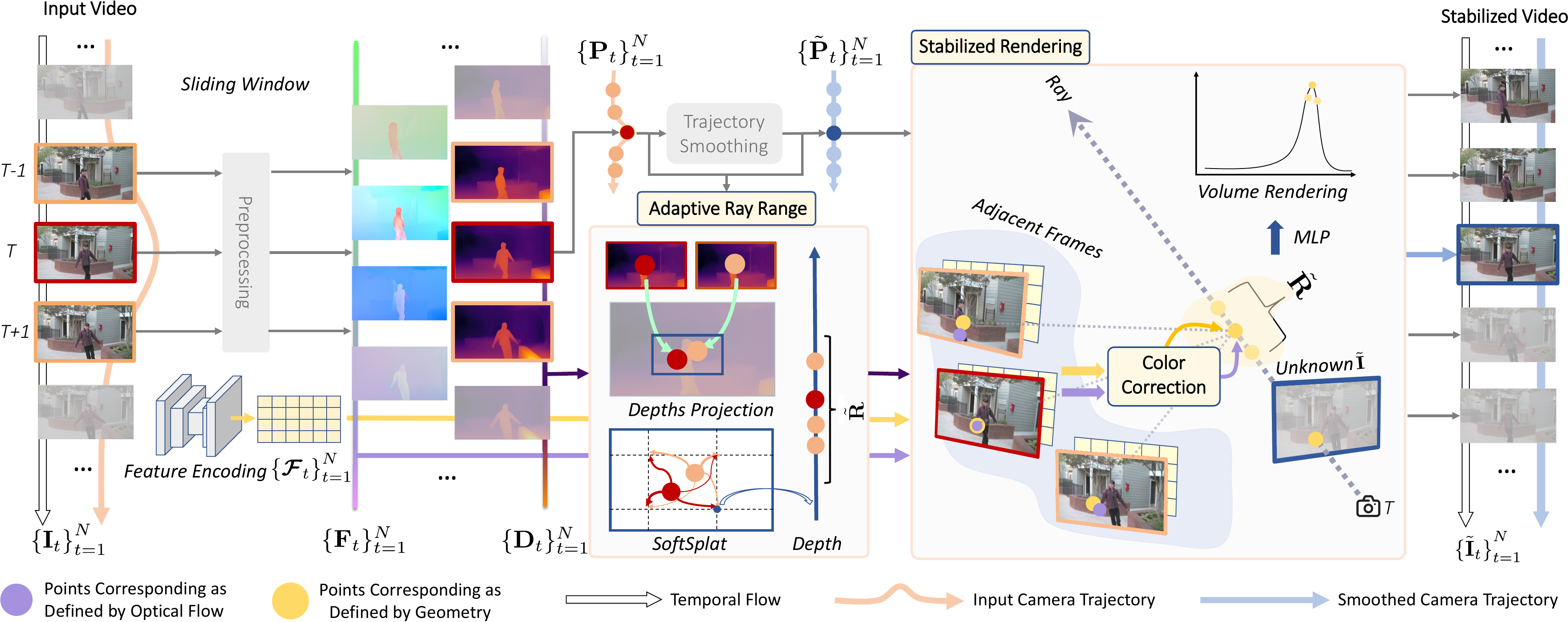}
    \caption{\textbf{Overview of our framework.}~(1)~Given input frames ${\{\mathbf{I}_t\}}_{t=1}^{N}$ with a shaky trajectory~${\{\mathbf{P}_t\}}_{t=1}^{N}$, our purpose lies in rendering stabilized video sequence ${\{\tilde{\mathbf{I}}_t\}}_{t=1}^{N}$ with smoothed trajectory ${\{\tilde{\mathbf{P}}_t\}}_{t=1}^{N}$. Here, the input trajectories ${\{\mathbf{P}_t\}}_{t=1}^{N}$ derive from preprocessing, while the smoothed trajectories ${\{\tilde{\mathbf{P}}_t\}}_{t=1}^{N}$ are generated using a Trajectory Smoothing module.
    ~(2)~In addition to ${\{\mathbf{P}_t\}}_{t=1}^{N}$, depth maps ${\{\mathbf{D}_t\}}_{t=1}^{N}$ and  optical flow $\{\mathbf{F}_{t}\}_{t=1}^{N}$ can be obtained during preprocessing. We aggregate ${\{\mathbf{D}_t\}}_{t=1}^{N}$ into the ray range ${\{\tilde{\mathbf{R}}_{t}\}}_{t=1}^{N}$ using the Adaptive Ray Range module.
    The ray range ${\{\tilde{\mathbf{R}}_{t}\}}_{t=1}^{N}$, along with $\{\mathbf{F}_{t}\}_{t=1}^{N}$ and the smoothed trajectory ${\{\tilde{\mathbf{P}}_t\}}_{t=1}^{N}$, serves as inputs to the Stabilized Rendering module. Conducting Stabilized Rendering, enhanced by the Color Correction module, we fuse the input frames ${\{\mathbf{I}_t\}}_{t=1}^{N}$ and their features ${\{\bm{\mathcal{F}}_t\}}_{t=1}^{N}$ to render the stabilized video sequence  ${\{\tilde{\mathbf{I}}_t\}}_{t=1}^{N}$.
    \vspace{-10pt}
    }

   \label{fig:pipeline}
\end{figure*}

\noindent{\textbf{Neural Rendering.}}~
As a significant work in view synthesis, NeRF\cite{lin2022enerf} attains photorealistic synthesized images through implicit volumetric representation and volume rendering. 
It combines multi-view information, leveraging $3$D geometric constraints and pixel-wise rendering to generate high-quality images without missing content from novel viewpoints. 
While NeRF-based methods~\cite{martin2021nerfwild,barron2021mip,barron2022mip360,muller2022instant,li2021nsff} produce impressive synthesized image quality, its limitation in per-scene training hampers its direct application in video stabilization.
Certain approaches~\cite{yu2021pixelnerf,wang2021ibrnet,lin2022enerf,xu2022pointner,trevithick2021grf,chen2021mvsnerf} strive to improve the generalization of NeRF, but they are not inherently well-suited for video stabilization tasks. 
Some recent methods~\cite{meuleman2023progressive,li2023dynibar} attempt to apply techniques in NeRF to stabilize videos, these approaches inherit the limitations of the vanilla NeRF, necessitating retraining for each specific scene. 
Inspired by generalized rendering technologies from IBRNet\cite{wang2021ibrnet} and ENeRF\cite{lin2022enerf}, which utilize multi-view images and associated features to predict radiance fields, 
we further propose the Stabilized Rendering. Stabilized Rendering, enhanced by the proposed Adaptive Ray Range module and Color Correction module, extends the volume rendering technique to video stabilization.

\section{Method} 
Our pipeline is shown in Fig.~\ref{fig:pipeline}. Given a shaky frame sequence ${\{\mathbf{I}_t\}}_{t=1}^{N}$ of length $N$, our objective is to generate a stabilized sequence ${\{\tilde{\mathbf{I}}_t\}}_{t=1}^{N}$. For preprocessing of ${\{\mathbf{I}_t\}}_{t=1}^{N}$, we estimate optical flow $\{\mathbf{F}_{t}\}_{t=1}^{N}$, depth maps ${\{\mathbf{D}_t\}}_{t=1}^{N}$, and camera trajectory ${\{\mathbf{P}_t\}}_{t=1}^{N}$ . With ${\{\mathbf{D}_t\}}_{t=1}^{N}$, ${\{\mathbf{P}_t\}}_{t=1}^{N}$ and smoothed camera trajectory ${\{\tilde{\mathbf{P}}_t\}}_{t=1}^{N}$ as input, the Adaptive Ray Range module aggregates multi-view depth maps into the ray ranges ${\{\tilde{\mathbf{R}}_{t}\}}_{t=1}^{N}$.  Guided by the ranges, Stabilized Rendering enhanced by the Color Correction module generates stabilized video sequence ${\{\tilde{\mathbf{I}}_t\}}_{t=1}^{N}$ through fusing the input frames ${\{\mathbf{I}_t\}}_{t=1}^{N}$ and feature maps ${\{\bm{\mathcal{F}}_t\}}_{t=1}^{N}$ obtained through feature extraction network.

We start with preprocessing a sequence of input frames ${\{\mathbf{I}_t\}}_{t=1}^{N}$ to estimate associated depth maps ${\{\mathbf{D}_t\}}_{t=1}^{N}$ and camera trajectory ${\{\mathbf{P}_t\}}_{t=1}^{N}$. These depth maps and camera poses are employed for camera trajectory smoothing. In our pursuit of consistent and smooth camera trajectories, we harness the flexibility of the Gaussian smoothing function: 
    ${\{\tilde{\mathbf{P}}_{t}\}}_{t=1}^{N} = \phi_{sm}({\{\mathbf{P}_t\}}_{t=1}^{N})$,
where $\phi_{sm}$ represents the Gaussian smoothing filter, offering adjustable parameters for both the smoothing window and stability. These parameters can be fine-tuned to meet specific requirements and constraints. In Sec.~\ref{Stabilized Rendering}, we elaborate on rendering a stabilized image with its neighboring frames through Stabilized Rendering.
Due to dynamic regions, the conventional $3$D-constraint-based rendering fails to adequately represent the geometry. 
Differing from the conventional rendering, Sec.~\ref{Adaptive Ray Range} introduces the utilization of depth priors to constrain the sampling range of spatial points around potential geometries, such as the area around the surface of objects. Additionally, in Sec.~\ref{Flow-based Color Correction}, we discuss refining projecting inaccuracy to ensure consistent local color intensities.

Stabilizing a video involves rendering a image sequence ${\{\tilde{\mathbf{I}}_t\}}_{t=1}^{N}$ with corresponding stabilized poses ${\{\tilde{\mathbf{P}}_{t}\}}_{t=1}^{N}$. In practice, we adopt a sliding window strategy for frame-by-frame rendering stabilized video. 
For clarity, we illustrate the rendering process with a single target camera pose $\tilde{\mathbf{P}}$ at the timestamp $T$ and its temporal neighborhood $\Omega_T$.

\subsection{Stabilized Rendering}
\label{Stabilized Rendering}

Stabilized Rendering is a multi-frame fusion module founded on epipolar constraints which
 fuses input images and feature maps to render a stable, uncropped video sequence. Considering a pixel $\tilde{\mathbf{x}}$ situated in the stabilized image $\tilde{\mathbf{I}}$ under a specific target camera pose $\tilde{\mathbf{P}}$, we sample $L$ spatial points sharing projection situation $\tilde{\mathbf{x}}$. These sampled points span depth ${\{\tilde{{d}}_i\}}_{i=1}^{L}$ distributed along the ray with sampling range, denoted as $\tilde{\mathbf{R}}(\tilde{\mathbf{x}})$. We project $\tilde{\mathbf{x}}$ at depth $\tilde{{d}}_i$ onto the neighboring input frames ${\{\mathbf{I}_t\}}_{t\in\Omega_T}$ at corresponding positions ${\{\mathbf{x}_{t}^{i}\}}_{t\in\Omega_T}$ by
\begin{equation}
\label{projection}
    \mathbf{x}_{t}^{i}=\mathbf{K}\mathbf{P}_t\tilde{\mathbf{P}}^{-1}\tilde{{d}}_i\mathbf{K}^{-1}\tilde{\mathbf{x}}\,,
\end{equation}
where $\mathbf{K}$ represents the camera intrinsic parameters shared by all frames in a video and $i\in (0,L]$. With the projected points ${\{\mathbf{x}_{t}^{i}\}}_{t\in\Omega_T}$, we aggregate features ${\{\bm{\mathcal{F}}_t(\mathbf{x}_t^{i})\}}_{t\in\Omega_T}$ in neighboring frames to predict the volume density $\sigma_i$ for the spatial point by
\begin{equation}
\label{density}
     \sigma_i=\phi_{mlp}({\{\bm{\mathcal{F}}_t(\mathbf{x}_t^{i})\}}_{t\in\Omega_T})\,, 
\end{equation}
where $\phi_{mlp}$ is a Multiple Layer Perceptron 
~(refer to Supp. for details). Eq.~\ref{density} is contingent upon the consistency among features. Specifically, if a sampled spatial point aligns with the ground geometry, the multi-view features of projected points would be similar. This condition establishes scene-independent geometric constraints. 
When considering the associated color $\mathbf{c}_i$, a conventional method is a linear combination for aggregation:
\begin{equation}
\label{color}
    \mathbf{c}_i=\sum_{t\in\Omega_T} \omega_{t-T}\mathbf{I}_t(\mathbf{x}_t^{i})\,, \sum_{t\in \Omega_T} \omega_{t-T}=1\,,
\end{equation}
where  $\omega_{t-T}$ represents adaptable parameters determined by the geometric characteristics, such as the volume density $\sigma_i$. Since the establishment of $\mathbf{c}_i$ solely relies on input frames,  it is training-free to accommodate unforeseen scenes. In volume rendering, the set ${\{\mathbf{c}_i,\sigma_i\}}_{i=1}^L$, describing spatial points along the same ray, determine the color intensity of $\tilde{\mathbf{x}}$ by
\begin{equation}
\label{volume rendering}
\begin{split}
    \tilde{\mathbf{I}}(\tilde{\mathbf{x}})&=\sum\limits_{i=1}^LA_i(1-exp(-\sigma_i))\mathbf{c}_i, \\
    A_i &=exp\left(-\sum\limits_{j=1}^{i-1}\sigma_i\right).
\end{split}
\end{equation}

In Stabilized Rendering, Eqs.~\ref{projection} imposes epipolar constraints on features and colors warped from multiple neighboring frames.
Eqs.~\ref{density}$~\&~$Eqs.~\ref{color} aggregate the multi-frame information into spatial descriptors~${\{\mathbf{c}_i,\sigma_i\}}_{i=1}^L$, and Eqs.~\ref{volume rendering} renders stabilized images utilizing these descriptors for each pixel. Epipolar constraints guarantee the structure preservation and per-pixel rendering guarantees full-frame generation. However, the effectiveness of the aforementioned process highly depends on the ray range $\tilde{\mathbf{R}}(\tilde{\mathbf{x}})$ guiding the sampling. If $\tilde{\mathbf{R}}(\tilde{\mathbf{x}})$ is not distributed near the surface of objects, the model may aggregate incorrect features into inferior descriptors and diminish rendering quality. The forthcoming section will introduce how to adaptively define the ray range $\tilde{\mathbf{R}}(\tilde{\mathbf{x}})$ to avoid the issue above.

\begin{figure}[t]
   \includegraphics[width=1\linewidth]{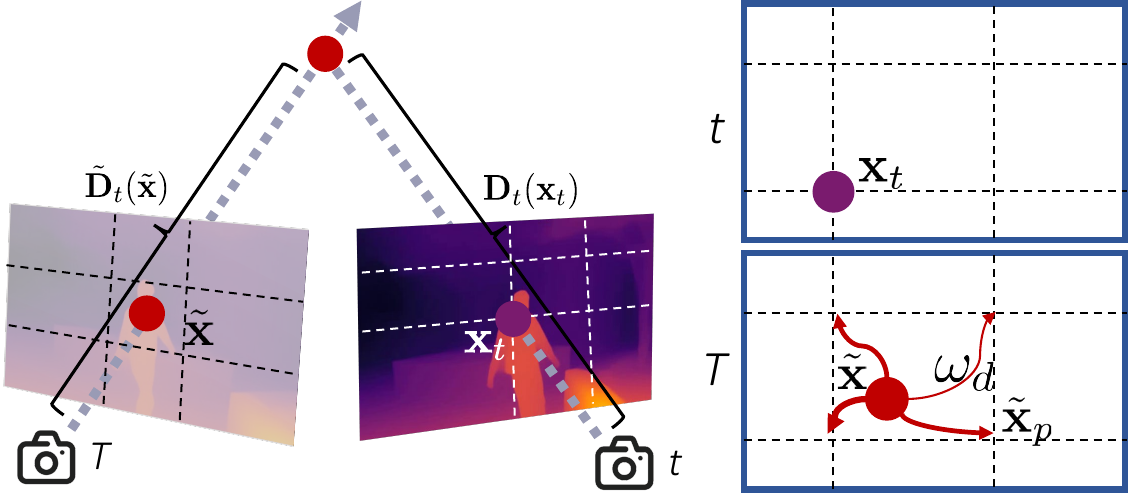}
   \caption{\textbf{Illustration of depth projection and splatting.}~Left:~The depth projection involve lifting a pixel $\mathbf{x}_t$ to $3$D space using the estimated depth $\mathbf{D}_t(\mathbf{x}_t)$ and projecting to the sub-pixel $\tilde{\mathbf{x}}$. The depth of $\tilde{\mathbf{x}}$ can be calculated and denoted as $\tilde{\mathbf{D}}_t(\tilde{\mathbf{x}})$. Right:~As $\tilde{\mathbf{x}}$ is not precisely projected onto a pixel coordinate, we convert its depth to adjacent pixels, e.g. $\tilde{\mathbf{x}}_p$, with a distance-associated weight $\omega_t$.}
   \label{fig:depth_proj}
\end{figure}

\subsection{Adaptive Ray Range}
\label{Adaptive Ray Range}

Eq.~\ref{volume rendering} of Stabilized Rendering highlights the dependence of the final color intensity of $\tilde{\mathbf{I}}(\tilde{\mathbf{x}})$ on the color $\mathbf{c}_i$ of the $3$D point where the ray hits the object for the first time. 
It indicates that ray ranges around the ground geometry for the sampling process will benefit scene representation.
A direct method to define the ray range entails treating the sequence of frames as a static scene: estimating the coarse geometry of each ray and rendering through spatial points sampled from re-defined fine ranges, such as~\cite{wang2021ibrnet,lin2022enerf}.
We argue that the effectiveness of the coarse-to-fine ray range relies on the geometry estimation grounded in epipolar constraints. 
However, dynamic regions, violating epipolar constraints, make the defined range unreliable.

To tackle this challenge, we turn to the task of depth estimation.  The depth model~\cite{lee2021deep3d} employs optical flow to impose constraints on dynamic scenes. As optical flow relies on feature matching rather than epipolar constraints, it matches points with features rather than epipolar constraints, showcasing insensitivity to dynamic regions. Consequently, the estimated depth maps derived from this depth model are less susceptible to interference from dynamic objects.
We propose to define an adaptive range with pre-estimated neighboring depth maps~${\{\mathbf{D}_t\}}_{t\in\Omega_T}$. 
In particular, we construct the range utilizing the mean and variance of aggregated depth maps from nearby frames.

\begin{figure}[t]
 
   \includegraphics[width=1\linewidth]{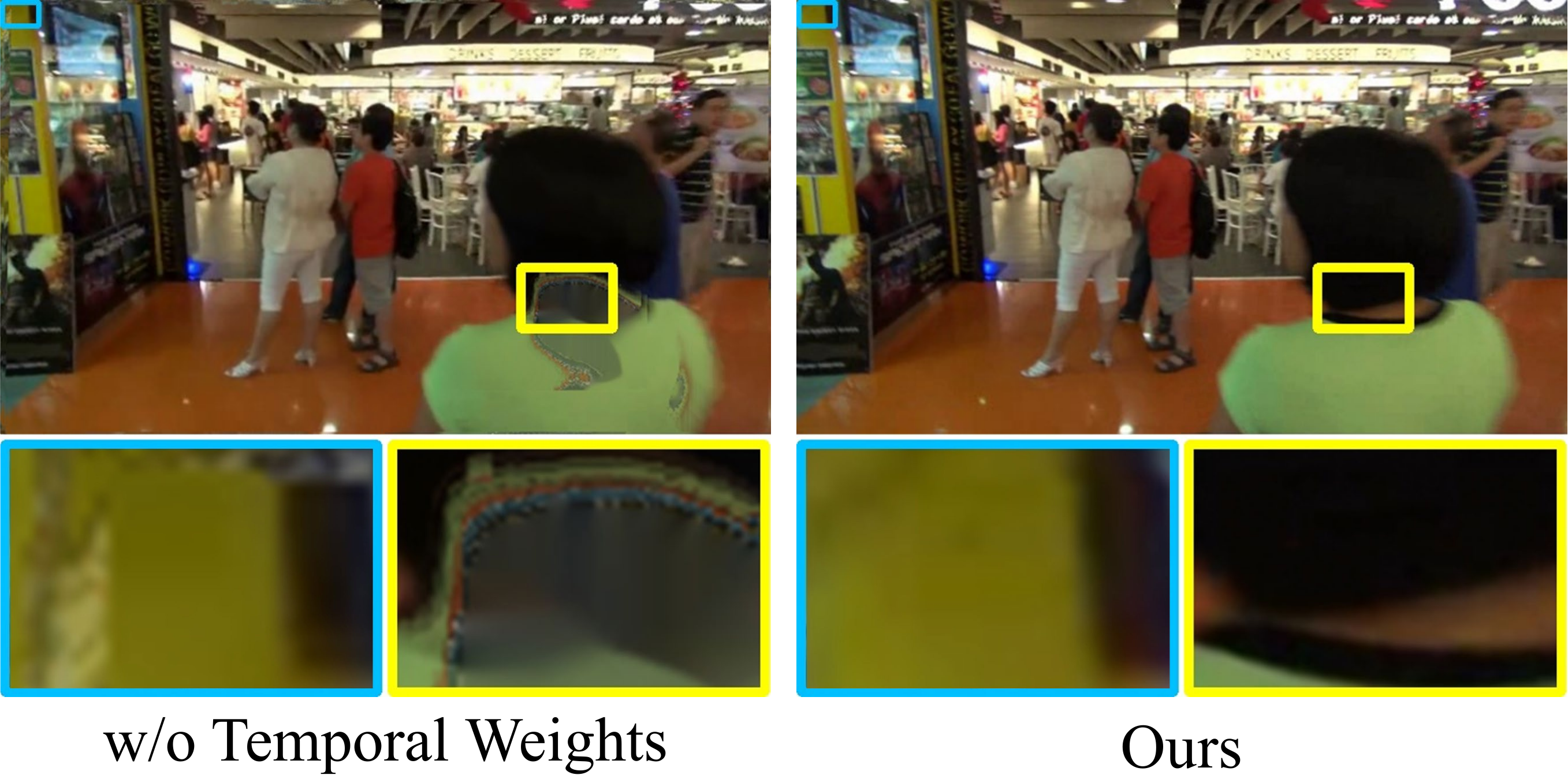}
   \caption{\textbf{The effect of temporal weights.} The introduction of temporal weights can mitigate distortion. }
   \label{fig:temporal_weights}

\end{figure}

As illustrated in the left part of Fig.~\ref{fig:depth_proj}, we project $\mathbf{x}_t$ in the neighboring frame with pose $\mathbf{P}_t$ at the depth $\mathbf{D}_t(\mathbf{x}_t)$ onto sub-pixel $\tilde{\mathbf{x}}$ of the stabilized frame with pose $\tilde{\mathbf{P}}$ according to the inverse of Eq.~\ref{projection}.
However, as sub-pixel $\tilde{\mathbf{x}}$ is not precisely projected onto a specific pixel coordinate, direct utilization of $\tilde{\mathbf{D}}_t(\tilde{\mathbf{x}})$ to estimate ray ranges for pixels is not feasible. To overcome this limitation, a splatting method~\cite{niklaus2020softsplat} is employed, as illustrated in the right part of Fig.~\ref{fig:depth_proj}, converting $\tilde{\mathbf{D}}_t(\tilde{\mathbf{x}})$ in the following manner:
\begin{equation}
    \tilde{\mathbf{D}}_t(\tilde{\mathbf{x}}_p)=\frac{\sum_i w_d \tilde{\mathbf{D}}_t(\tilde{\mathbf{x}}_i)}{\sum_i w_d}, w_d=\prod (\mathbf{1}-|\tilde{\mathbf{x}}_p-\tilde{\mathbf{x}}_i|)\,,
\end{equation}

where $\tilde{\mathbf{x}}_p$ is a pixel and $\tilde{\mathbf{x}}_i$ is the $i$-th sub-pixel $\tilde{\mathbf{x}}$ around $\tilde{\mathbf{x}}_p$ satisfying the condition $|\tilde{\mathbf{x}}_p-\tilde{\mathbf{x}}_i|\in (0,1)^2$, $\prod(\cdot)$ suggests an element-wise multiplication in a vector, and $\omega_d$ is distance-associated weights.

Given ${\{\mathbf{D}_t\}}_{t\in\Omega_T}$, we obtain corresponding ${\{\tilde{\mathbf{D}}_t\}}_{t\in\Omega_T}$ on the stabilized frame through the project-splat process above. An intuitive approach involves directly calculating the mean $\mathbf{M}$, variance $\mathbf{S}$, and determining the sampling ray range as ${\mathbf{R}} = \left[{\mathbf{M}}-{\mathbf{S}},{\mathbf{M}}+{\mathbf{S}}\right].$
However, in the aforementioned depth project-splat process, depth maps further from the timestamp T are less reliable. Treating all depth maps equally can result in an inaccurate sampling ray range ${\mathbf{R}}$, leading to a decrease in the image quality~(the left part of Fig.~\ref{fig:temporal_weights}).
\begin{equation}
    \tilde{\mathbf{M}} = \sum\limits_{t\in\Omega_T}\omega_t\tilde{\mathbf{D}}_t\,,
    \tilde{S}=\sqrt{\sum\limits_{t\in\Omega_T}\omega_t(\tilde{\mathbf{D}}-\tilde{\mathbf{M}})^2},
\end{equation}
where $\omega_t$ is the temporal weighting coefficient, assigning a higher weight to the frame closer to the stabilized frame temporally and vice versa, as defined by
\begin{equation}
\label{weight}
    \omega_{t} =\frac{e^{\lambda(t-T)}}{\sum\limits_{t\in \Omega_T}e^{\lambda(t-T)}}\,,
\end{equation}
where $\lambda$ is a hyperparameter. Subsequently, ray ranges for the stabilized frame are denoted as $\tilde{\mathbf{R}} = \left[\tilde{\mathbf{M}}-\tilde{\mathbf{S}},\tilde{\mathbf{M}}+\tilde{\mathbf{S}}\right].$ and can be employed for sampling L points along each ray during the rendering process. 
As illustrated in the right part of Fig.~\ref{fig:temporal_weights}, the Adaptive Ray Range module with temporal weighted ranges yields more favorable rendering results.

The Adaptive Ray Range module provides a ray range $\tilde{\mathbf{R}}$ around the ground geometry guiding points sampling and benefiting volume density $\sigma_i$ prediction. Although the guidance of $\tilde{\mathbf{R}}$ mitigates the interference of dynamic objects, the challenge of dynamic objects goes beyond this. According to Eq.~\ref{volume rendering}, the color intensity $\mathbf{c}_i$ is another factor influencing rendering quality and affected by dynamic regions as well.

\subsection{Color Correction}
\label{Flow-based Color Correction}
\begin{figure}[t]
   \includegraphics[width=1\linewidth]{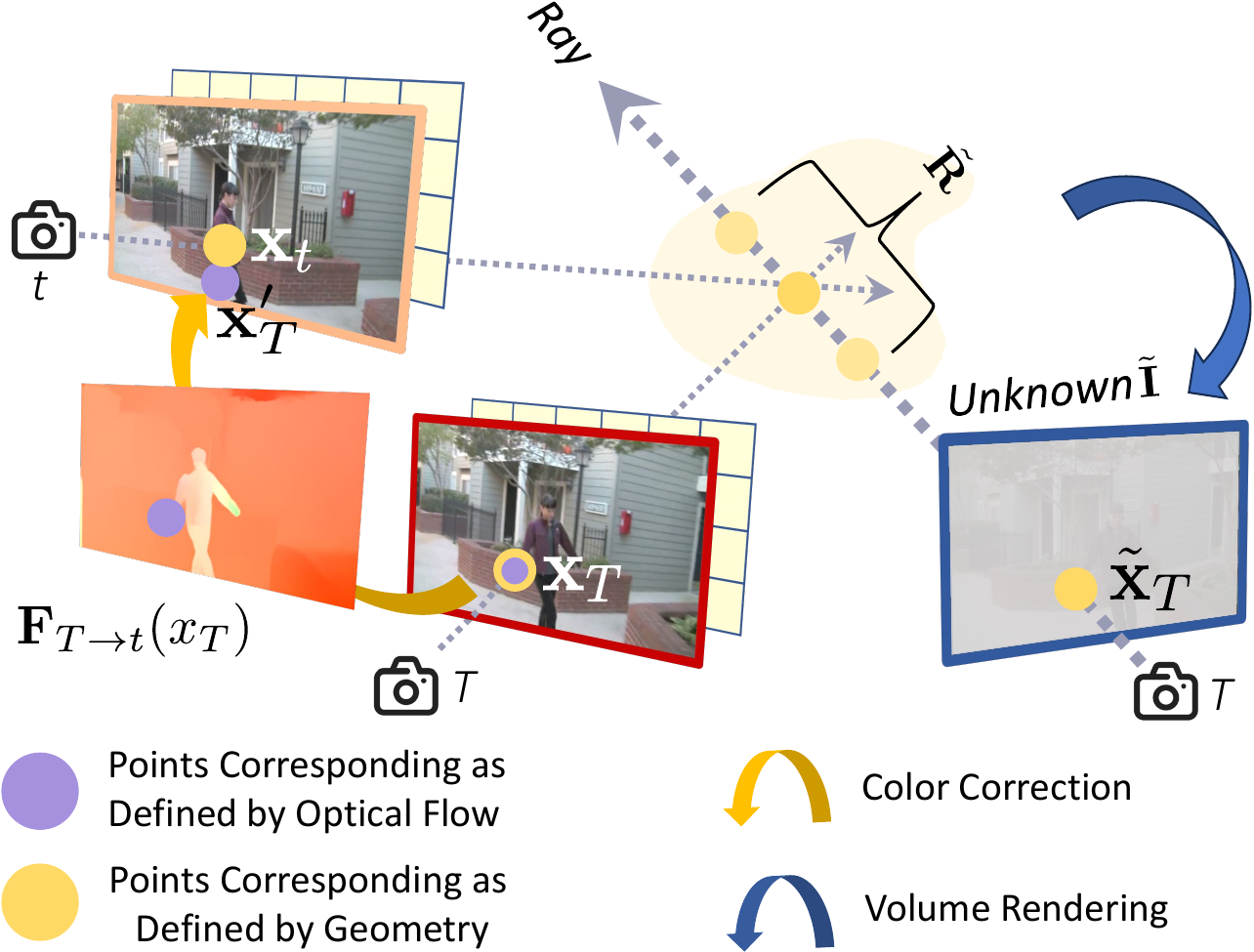}
      \caption{\textbf{Illustration of Color Correction module.}~ 
      Firstly, we project a pixel $\tilde{\mathbf{x}}_{T}$ from the target stabilized frame onto corresponding ${\mathbf{x}}_{T}$ of the input frame at the same timestamp $T$. Secondly, we obtain feature matching of ${\mathbf{x}}_{T}$ in the input frame at timestamps $t$ using optical flow $\mathbf{F}_{{T}\rightarrow t}(\mathbf{x}_{T})$. As geometric constraints alone are insufficient for modeling dynamic regions, we aggregate precise color by correcting the geometric projected position $\mathbf{x}_t$ to the optical-flow refined position $\mathbf{x}'_{t}$.
      } 
   \label{fig:correction}
\end{figure}

\begin{table*}\small
    \centering
    \begin{tabular}{@{}lcccclccclccc@{}}
\toprule
\multirow{2}{*}{Method}&&\multicolumn{3}{c}{NUS dataset}& &\multicolumn{3}{c}{Selfie dataset}& &\multicolumn{3}{c}{DeepStab dataset}\\ 
\cmidrule(lr){3-5} \cmidrule(lr){7-9} \cmidrule(l){11-13}
  &  &C$\uparrow$ &D$\uparrow$ &S$\uparrow$ & &C$\uparrow$ &D$\uparrow$ &S$\uparrow$ & &C$\uparrow$ &D$\uparrow$ &S$\uparrow$\\
\midrule

Grundmann~\etal~\cite{grundmann2011l1} &$2$D &0.71  &0.76  &0.82 & &0.75 &0.81 &0.83 & &0.77  &0.87  &0.84   \\
Bundle~\cite{liu2013bundle}    &$2$D    &0.81  &0.78  &0.82 & &0.74 &0.82&0.80 & &0.80  &0.90  &\underline{0.85}\\
Yu and Ramamoorthi~\cite{yu2020stable} &$2$D &\underline{0.85}  &0.81  &0.86 & &\underline{0.83} &0.79&0.86 & &\underline{0.87}  &0.92  &0.82\\
DIFRINT~\cite{choi2020difrint}   &$2$D       &\textbf{1.00}  &0.87  &0.84 & &\textbf{1.00} &0.78 &0.84 & &\textbf{1.00}  &0.91  &0.78\\
FuSta~\cite{liu2021fusta}        &$2$D      &\textbf{1.00}  &0.87  &0.86 & &\textbf{1.00} &0.83 &\underline{0.87} & &\textbf{1.00}  &0.92  &0.82\\
Zhao~\etal~\cite{zhao2023fast}    &$2$D     &\textbf{1.00}  &\underline{0.90}  &\underline{0.87} & &\textbf{1.00} &\underline{0.87}&\underline{0.87} & &\textbf{1.00}  &\underline{0.94}  &0.84\\ \midrule
Deep$3$D~\cite{lee2021deep3d}    &$3$D     &0.66  &\underline{0.90}  &\textbf{0.94} & &0.35 &0.70 &\textbf{0.95} & &0.75  &\textbf{0.98}  &\textbf{0.92}\\
Ours                        &$3$D    &\textbf{1.00}  &\textbf{0.91}  &\textbf{0.94} & &\textbf{1.00} &\textbf{0.92} &\textbf{0.95} & &\textbf{1.00}  &\textbf{0.98}  &\textbf{0.92}\\
    \bottomrule
    \end{tabular}
    \caption{\textbf{Quantitative results on the NUS~\cite{liu2013bundle}, the Selfie~\cite{yu2018selfie}, and the DeepStab~\cite{wang2019deepstab} datasets.} 
    We evaluate our method against baselines using three standard metrics: Cropping Ratio(C), Distortion Value(D), Stability Score(S). The best results are \textbf{bolded} and second-best results are hightlighted by {\underline{underline}}.}
    \label{tab:quantitative}
\end{table*}

Color intensity, denoted as $\mathbf{c}_i$, exhibits a strong dependence on geometric constraints, akin to volume density $\sigma_i$.
However, density is predicted from the feature maps with their receptive fields, thereby exhibiting a certain tolerance to projection inaccuracy. In contrast, color intensity is derived from the linear combination of colors warped from multiple views, accentuating the sensitivity of colors to projection inaccuracy. 
Despite the Adaptive Ray Range module offers a correction for projection with geometric constraints, it is inadequate for accurate color aggregation~(refer to the experiments).
Rather than solely concentrating on refining geometric constraints, we propose to assist these constraints with optical flow.
Optical flow, relying on feature similarities, matches pixels with similar textures containing color information.
It implies that utilizing optical flow to refine the projection can enhance color accuracy.

Specifically, we focus on the input frame at $T$, which adheres to epipolar constraints with the target stabilized frame at $T$. As shown in Fig.~\ref{fig:correction}, we employ $\mathbf{I}_{T}$ as a reference to correct the projection points on the neighboring frame $\mathbf{I}_{t}$ with optical flow.
According to Eq.~\ref{projection}, we project a point $\tilde{\mathbf{x}}_{T}$ from the stabilized pose $\tilde{\mathbf{P}}_{T}$ onto the ${\mathbf{x}}_{T}$ of $\mathbf{P}_{T}$, the flow-associated points ${\mathbf{x}}'_{t}$ can be expressed as
\begin{equation}
    \mathbf{x}'_{t} = \mathbf{x}_{T}+\mathbf{F}_{{T}\rightarrow t}(\mathbf{x}_{T}),
\end{equation}
where $\mathbf{F}_{{T}\rightarrow t}$ represents the optical flow from $\mathbf{I}_{T}$ to $\mathbf{I}_t$. By applying the same procedure to frames in the temporal neighborhood $\Omega_T$, we substitute the $\mathbf{x}_t$ in Eq.~\ref{color} with $\mathbf{x}'_{t}$.

\subsection{Implementation Details}
In our implementations, a pre-trained model from Deep$3$D~\cite{lee2021deep3d} is employed to generate depth prior for the Adaptive Ray Range module and optical flow for Color Correction. Frames neighboring the timestamp T are symmetrically distributed, and the length of the set $\Omega_T$ is fixed to 13. For the Adaptive Ray Range module, the temporal weighting coefficient $\omega_i$ is calculated with $\lambda=0.5$, and we choose $L=3$ for uniform spatial points sampling along each ray.

\noindent\textbf{Loss function.}~During training, we sample rays on all images randomly and minimize the mean squared error between the rendered color and corresponding ground truth: 
\begin{equation}
    \mathcal{L}=\sum\limits_{\mathbf x \in \mathcal{X}}\left|\left|\tilde{\mathbf{I}}(\mathbf{x})-\mathbf{I}_{gt}(\mathbf{x})\right|\right|^2_2,
\end{equation}
where $\mathbf{I}_{gt}$ is the corresponding ground truth and $\mathcal{X}$ is the set of pixels sampled from all images in each training batch.

\noindent\textbf{Training details.}~
We follow the training setting of IBRNet~\cite{wang2021ibrnet} to train our model on LLFF~\cite{mildenhall2019llff} and IBRNetCollected~\cite{wang2021ibrnet} including high-quality natural images with accurate camera poses. Our model is trained on an RTX3090 GPU using the Adam optimizer\cite{kingma2015adam}. We set the base learning rates for the feature extraction network and MLP to $1e^{-3}$ and $5e^{-4}$, respectively, which decay exponentially throughout the optimization process. Typically, the model converges after approximately 200k iterations, and the entire training process takes about a day to complete.

\begin{figure*}
    \setlength{\tabcolsep}{0.7pt}
    \small
    \begin{tabular}{cccc}
        \begin{minipage}[b]{0.5145\columnwidth}
        \includegraphics[width=1\linewidth]{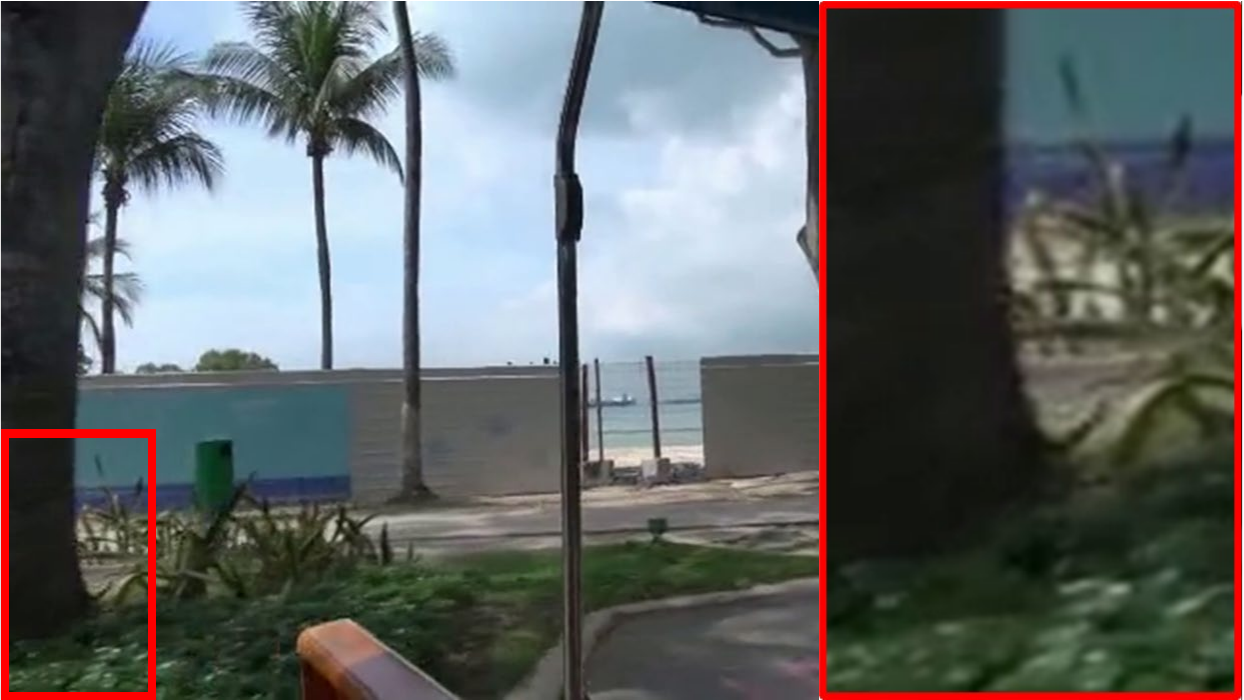}
        \end{minipage}
        &  
        \begin{minipage}[b]{0.5145\columnwidth}
        \includegraphics[width=1\linewidth]{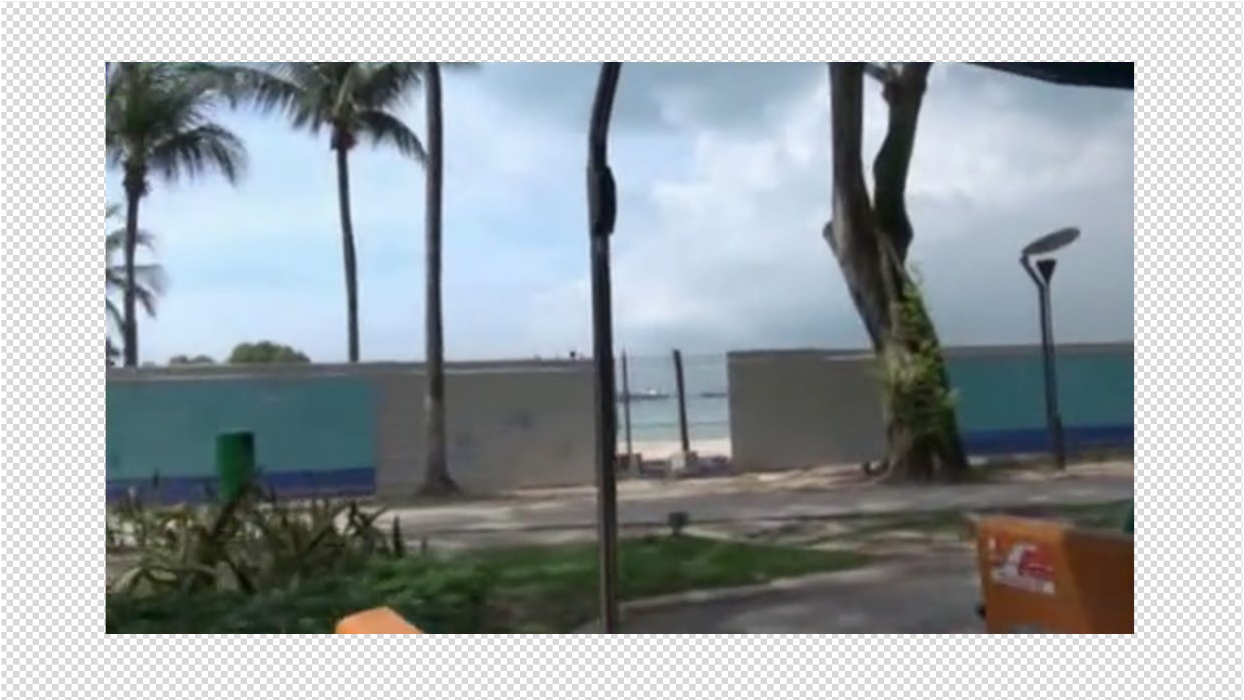}
        \end{minipage}
        &  
        \begin{minipage}[b]{0.5145\columnwidth}
        \includegraphics[width=1\linewidth]{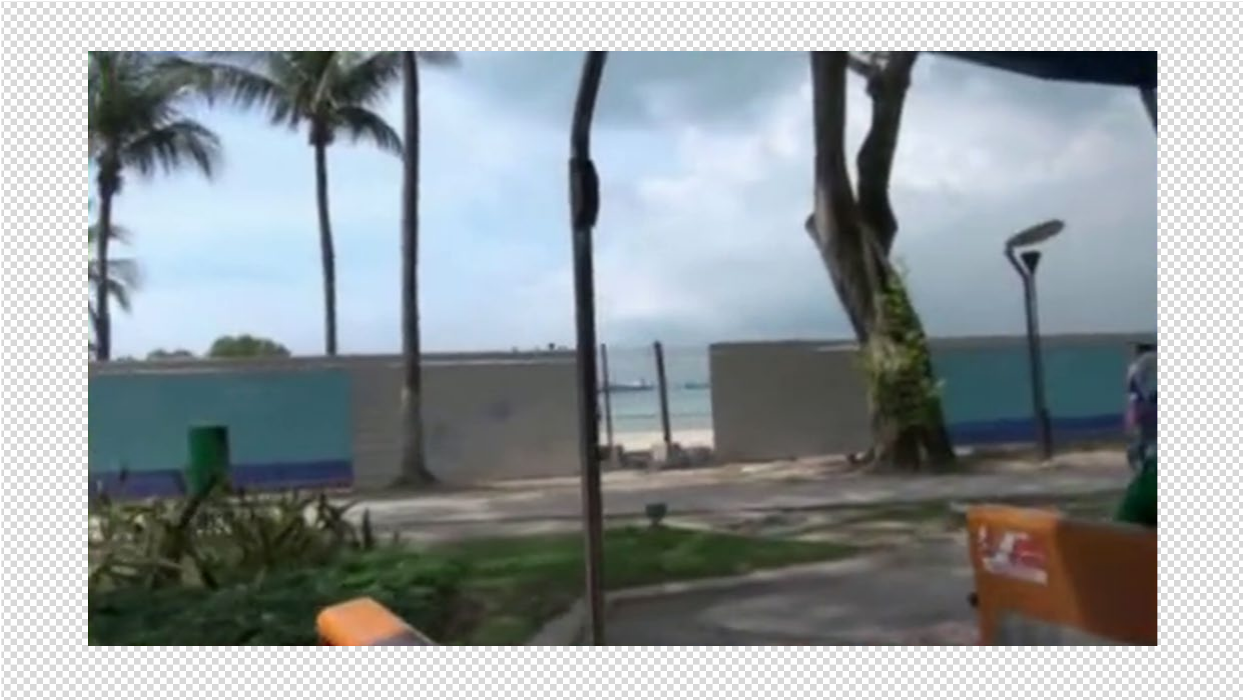}
        \end{minipage}
        &  
        \begin{minipage}[b]{0.5145\columnwidth}
        \includegraphics[width=1\linewidth]{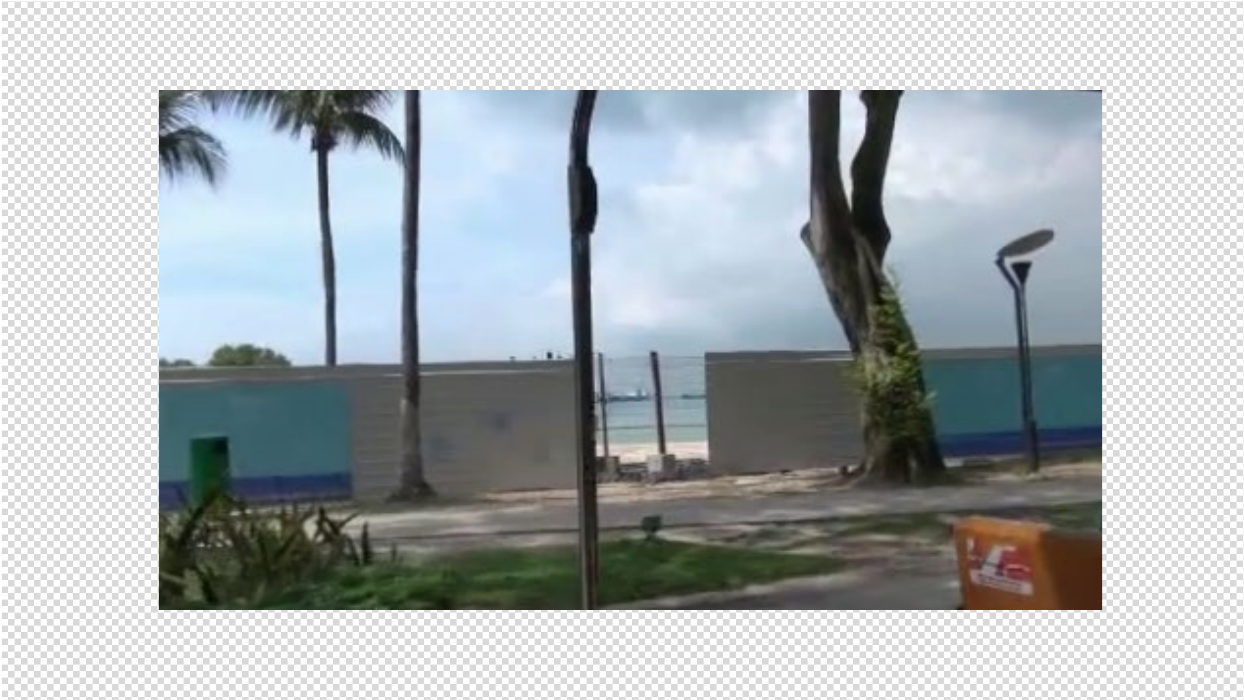}
        \end{minipage}
        \\
        
        Input  & Bundle~\cite{liu2013bundle}  & Yu and Ramamoorthi~\cite{yu2020stable}  & Deep$3$D\cite{lee2021deep3d}  \\

        \vspace{-8pt}
        
        \\
        \begin{minipage}[b]{0.5145\columnwidth}
        \includegraphics[width=1\linewidth]{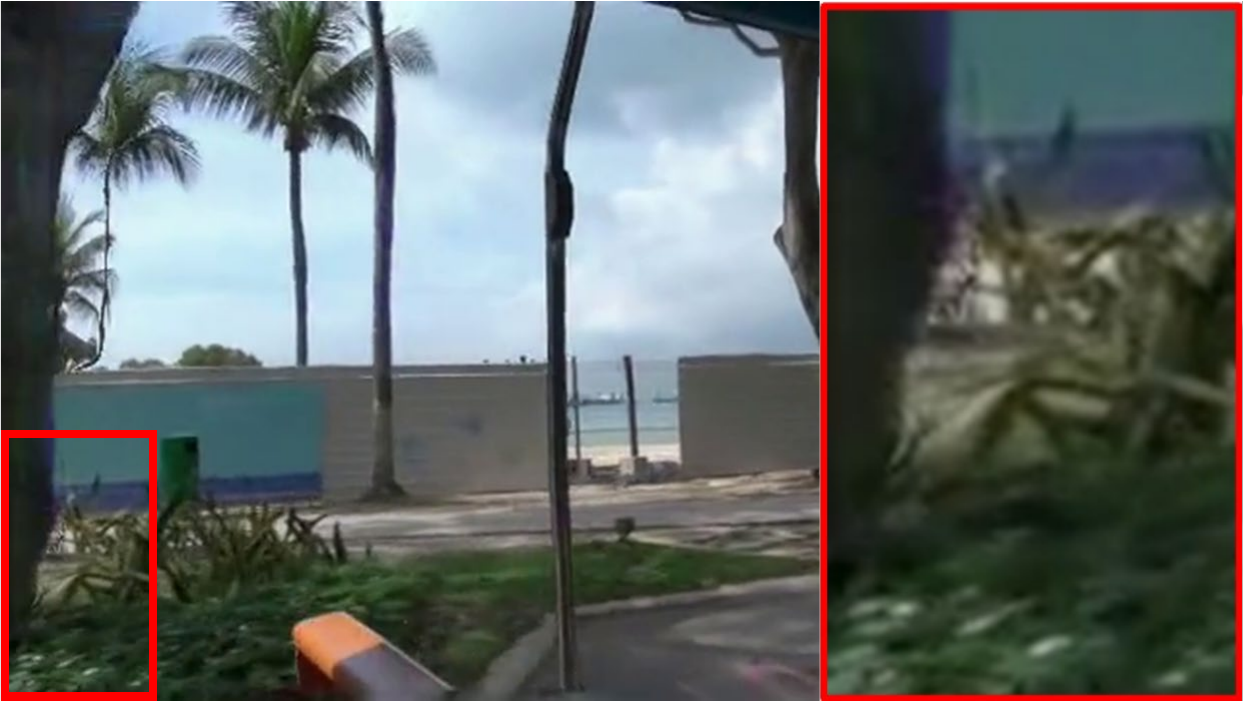}
        \end{minipage}
        &  
        \begin{minipage}[b]{0.5145\columnwidth}
        \includegraphics[width=1\linewidth]{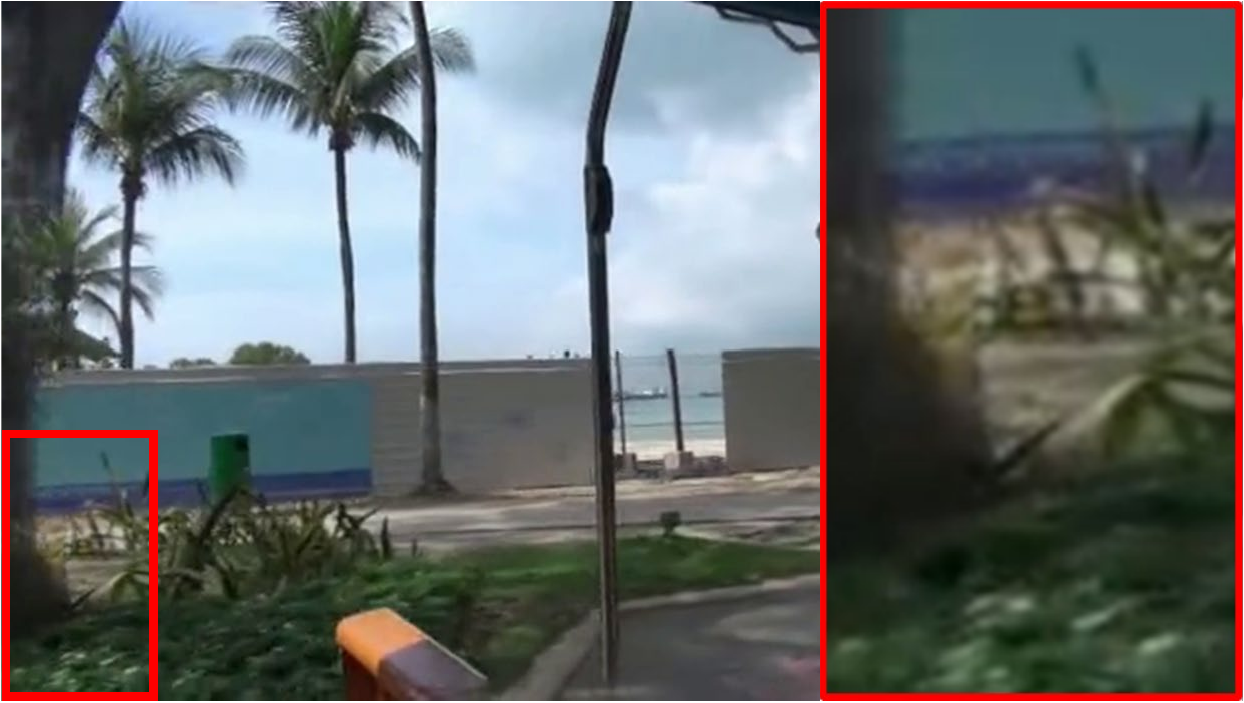}
        \end{minipage}
        &  
        \begin{minipage}[b]{0.5145\columnwidth}
        \includegraphics[width=1\linewidth]{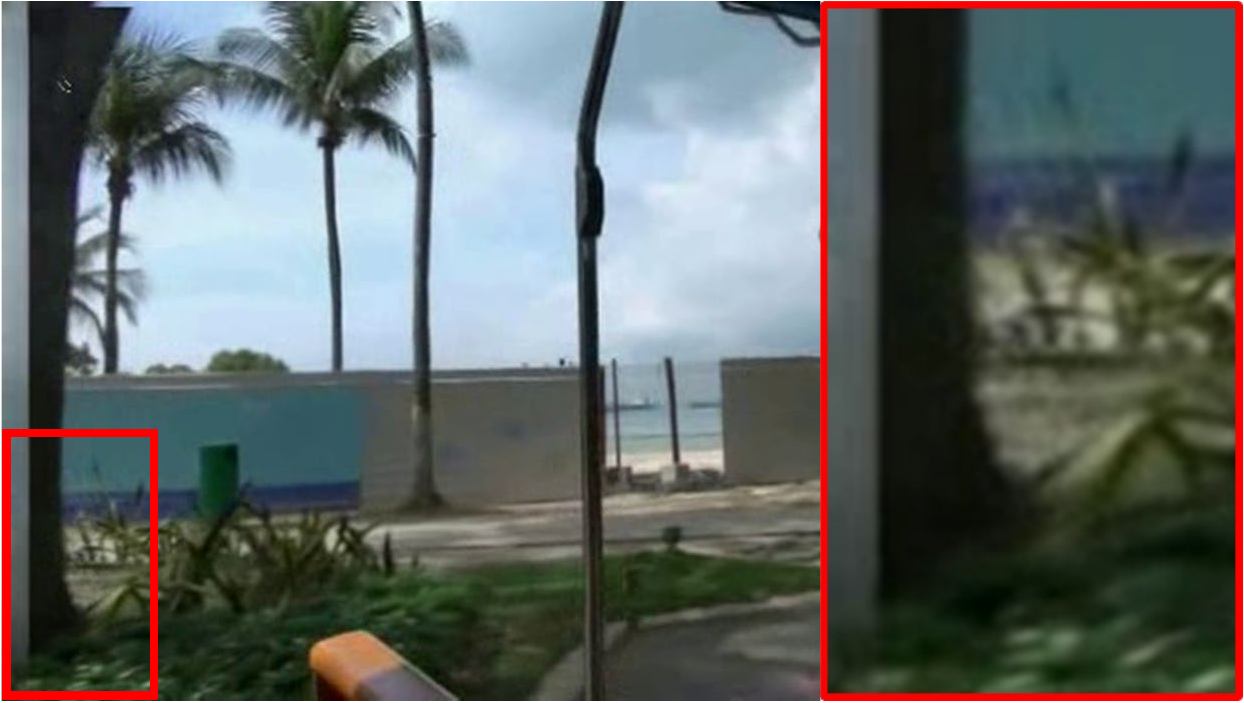}
        \end{minipage}
        &  
        \begin{minipage}[b]{0.5145\columnwidth}
        \includegraphics[width=1\linewidth]{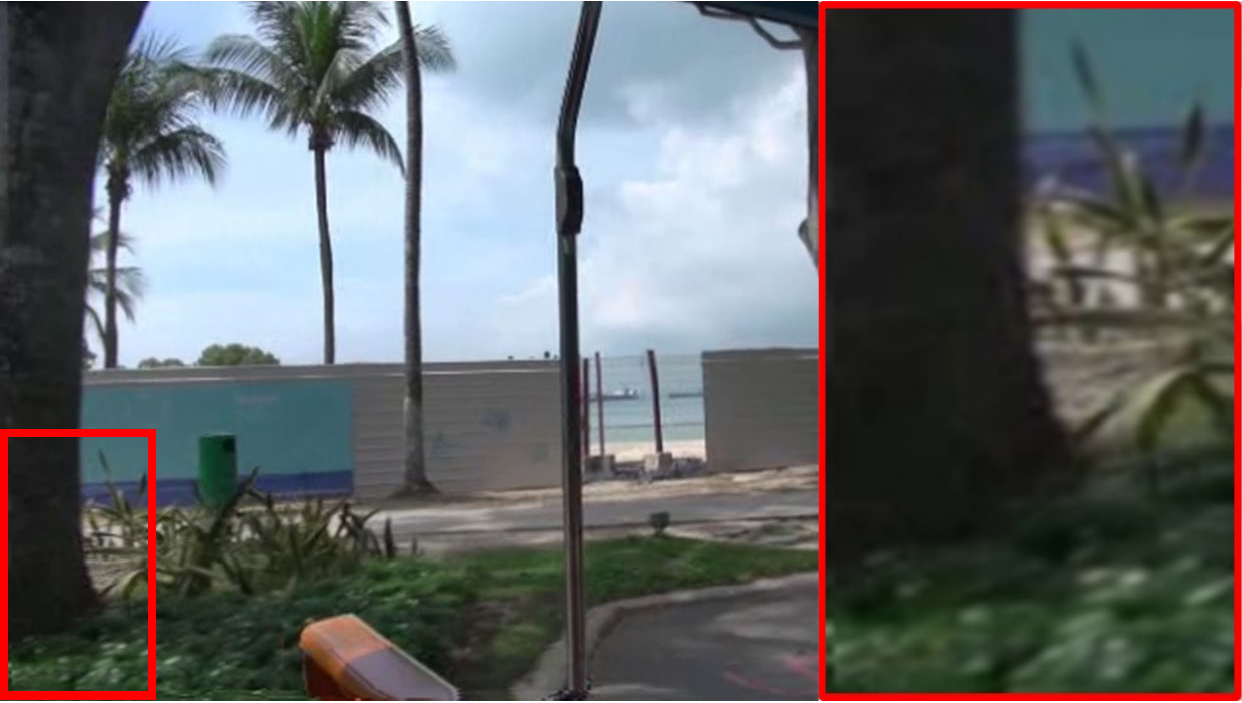}
        \end{minipage}

        \\ 
        DIFRINT~\cite{choi2020difrint}  & FuSta~\cite{liu2021fusta}  & Zhao \etal~\cite{zhao2023fast}  & Ours  \\
    \end{tabular}
    \caption{\textbf{Visual comparison of different methods.}~Contrasting with the baselines in the first row, our method successfully accomplishes full-frame generation. In the second row, while these baselines achieve full-frame generation, they fall short in preserving structure; for instance, in the bottom-left region, the tree trunks are missing in their stabilized images. Please refer to our supplementary material for video comparisons with baselines.} 
   \label{fig:visual}
   
\end{figure*}

\section{Experiments}

\subsection{Quantitative Evaluation}

\noindent\textbf{Baselines.}~We choose various video stabilization algorithms as the baselines, including Grundmann~\etal~\cite{grundmann2011l1}, Liu~\etal~\cite{liu2013bundle}, Wang~\etal~\cite{wang2019deepstab}, Yu and Ramamoorthi~\cite{yu2019robust, yu2020stable}, DIFRINT~\cite{choi2020difrint}, FuSta~\cite{liu2021fusta}, Zhao~\etal~\cite{zhao2023fast}, and Deep$3$D~\cite{lee2021deep3d}. For comparisons, we use the official-provided videos or videos generated by official implementations with default parameters or pre-trained models.

\vspace{5pt}
\noindent\textbf{Datasets.}
We choose three datasets with different characteristics for evaluations:~(1)~The NUS~\cite{liu2013bundle} dataset comprises 144 videos, categorized into six different scenes:~Regular, Running, Crowd, Parallax, QuickRotation, and Running, (2)~the Selfie dataset~\cite{yu2018selfie} contains 33 video clips featuring frontal faces with large camera motion, (3)~and the DeepStab dataset~\cite{wang2019deepstab} includes 61 high-definition videos. 

\vspace{5pt}
\noindent\textbf{Metrics.}~We assess the performance of the stabilizers using three standard metrics widely employed in previous methods~\cite{choi2020difrint, liu2021fusta, liu2013bundle, yu2019robust, yu2020stable}: (1)~\textit{Cropping Ratio}: This metric measures the remaining image area after cropping the non-content pixels. (2)~\textit{Distortion Value}: This metric quantifies the anisotropic scaling of the homography matrix between the input and output frames. (3)~\textit{Stability Score}: This metric assesses the stability of the stabilized video by assessing the ratio of low-frequency motion energy to the total energy. All three metrics range from 0 to 1, with higher values indicating better performance.

\vspace{5pt}
\noindent\textbf{Results on the NUS dataset.}
Our evaluation on the NUS dataset~\cite{liu2013bundle} is detailed on the left side of Table~\ref{tab:quantitative}, where our stabilization method excels notably in both stability and distortion reduction when compared to 2D-based methods. This success is attributed to our accuracy in constructing camera trajectories and geometry. 
In contrast to $3$D methods, our approach stands out by leveraging information from multiple input frames, achieving an average cropping ratio of 1. This indicates the effectiveness of our method in full-frame generation across the diverse scenes in the NUS dataset, which is widely acknowledged as a robust benchmark for video stabilization algorithms.

\noindent\textbf{Results on the Selfie dataset.}~
We present the results on the Selfie dataset~\cite{yu2018selfie} in the middle of Table~\ref{tab:quantitative}. It's crucial to highlight that this dataset is characterized by large camera motions and extensive dynamic regions, posing challenges for video stabilization algorithms.
Observing the results, a decrease is evident for most algorithms compared to their performance on the NUS dataset. Traditional $3$D methods, in particular, experience a significant decline. In contrast, our method consistently delivers the best performance on the Selfie dataset. The performance shows the effectiveness of our algorithm in handling extreme scenes.

\noindent\textbf{Results on the DeepStab dataset.}~
The right side of Table~\ref{tab:quantitative} showcases the average scores on the DeepStab dataset~\cite{wang2019deepstab}. Notably, the videos in this dataset are of higher resolution than NUS and Selfie, specifically 720p, aligning with the common resolutions of modern devices.
Despite the high distortion values across all stabilizers due to the simplicity of this dataset, our approach consistently demonstrates superior performance. This result suggests that our method is well-suited for handling high-definition videos, further emphasizing its applicability for contemporary video stabilization challenges.

\subsection{Qualitative Analysis}
\label{sec:Qualitative Analysis}
Visual comparisons of our method and state-of-the-art stabilizers is shown in Fig.~\ref{fig:visual}. Many methods~\cite{liu2013bundle,yu2020stable,lee2021deep3d} apply aggressive cropping, as evident from the grey checkerboard regions. Comparing the bottom-left region of each image in Fig.~\ref{fig:visual} below with the top-left input, it's clear that our method suffers from fewer visual artifacts. 

\begin{figure}[t]
   \includegraphics[width=1\linewidth]{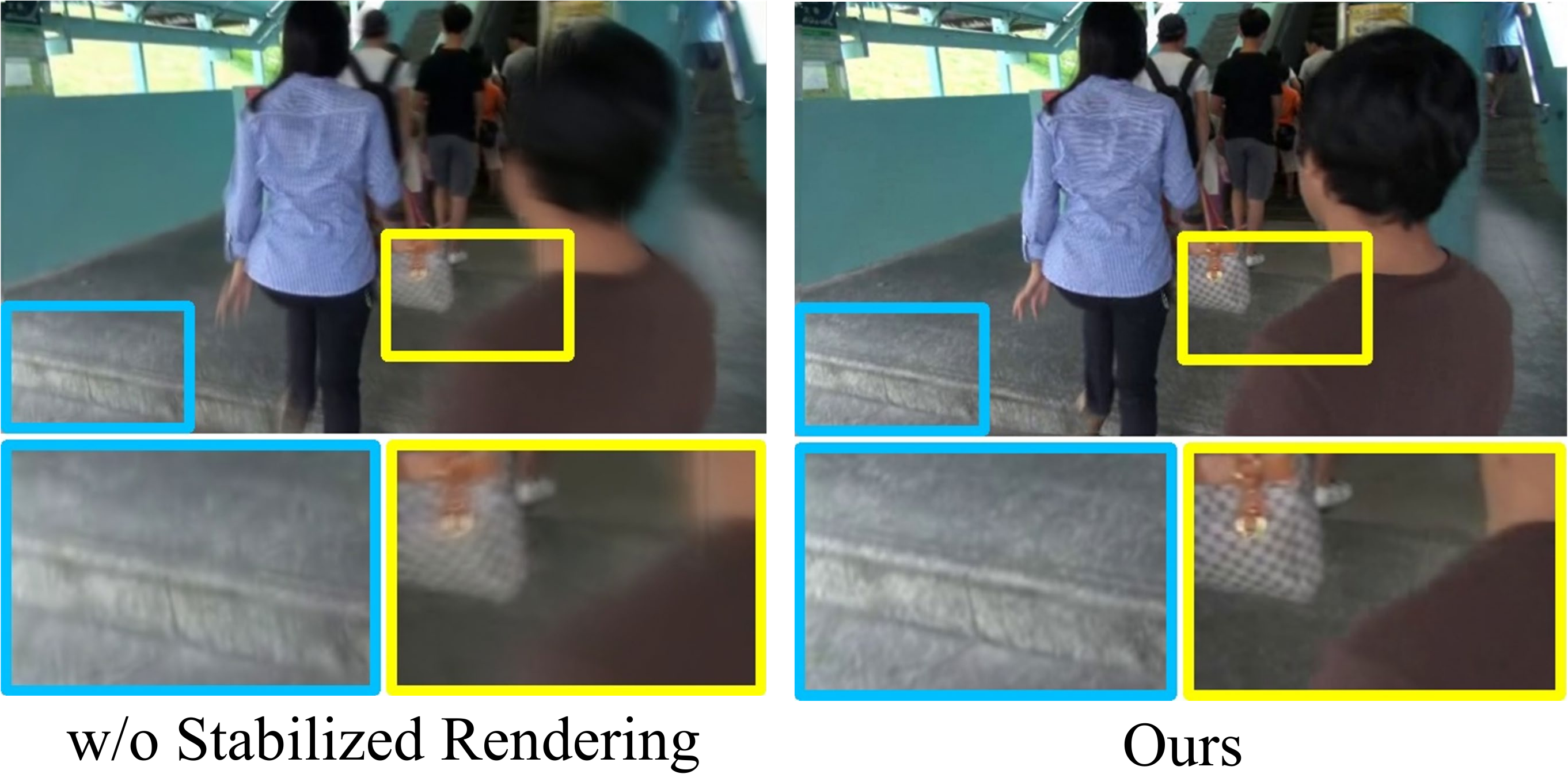}
   \caption{\textbf{Quilitative ablation of Stabilized Fusion.}~Absence of Stabilized Fusion results in noticeable blurs in both static and dynamic regions.} 
   \label{fig:stabilized_rendering}
\end{figure}

\section{Ablation Study}
\label{sec:Ablation Study}
We conduct ablation studies to analyze the effectiveness of the proposed modules, including \textbf{S}tabilized \textbf{R}endering~(SR), the \textbf{A}daptive \textbf{R}ay \textbf{R}ange module (ARR), and \textbf{C}olor \textbf{C}orrection module (CC). Our evaluations focus on the Crowd scene within the NUS dataset~\cite{liu2013bundle}, chosen for its dynamic objects and diverse scenes. 
We choose Distortion values and PSNR as evaluation metrics.
Distortion Value measures the pose-independent structure quality of images with stabilized poses. 
Additionally, PSNR is employed to evaluate the pixel-level performance of our model in rendering image details. As real images with stabilized poses are unavailable, we render images with the input pose to derive PSNR.

\noindent\textbf{Why needs Stabilized Rendering.}~
We conduct experiments to demonstrate the necessity of SR, which fuses features and colors in $3$D space. 
One straightforward strategy replacing SR for fusing multiple frames is image blending. It warps nearby frames into the stabilized view and averages these images. However, as illustrated in the left part of Fig.~\ref{fig:stabilized_rendering}, image blending leads to noticeable blur in both static regions~(the stairs) and dynamic regions~(the handbag and the shoulder). 
Comparing Row~4 and Row~3 in Table~\ref{tab:quantitative}, the notable decreases in distortion value and PSNR align with the observation in Fig.~\ref{fig:stabilized_rendering}. 
It demonstrates SR, our $3$D multi-frame fusion module using volume rendering,  can enhance the structural quality of stabilized images.
\begin{figure}[t]
   \includegraphics[width=1\linewidth]{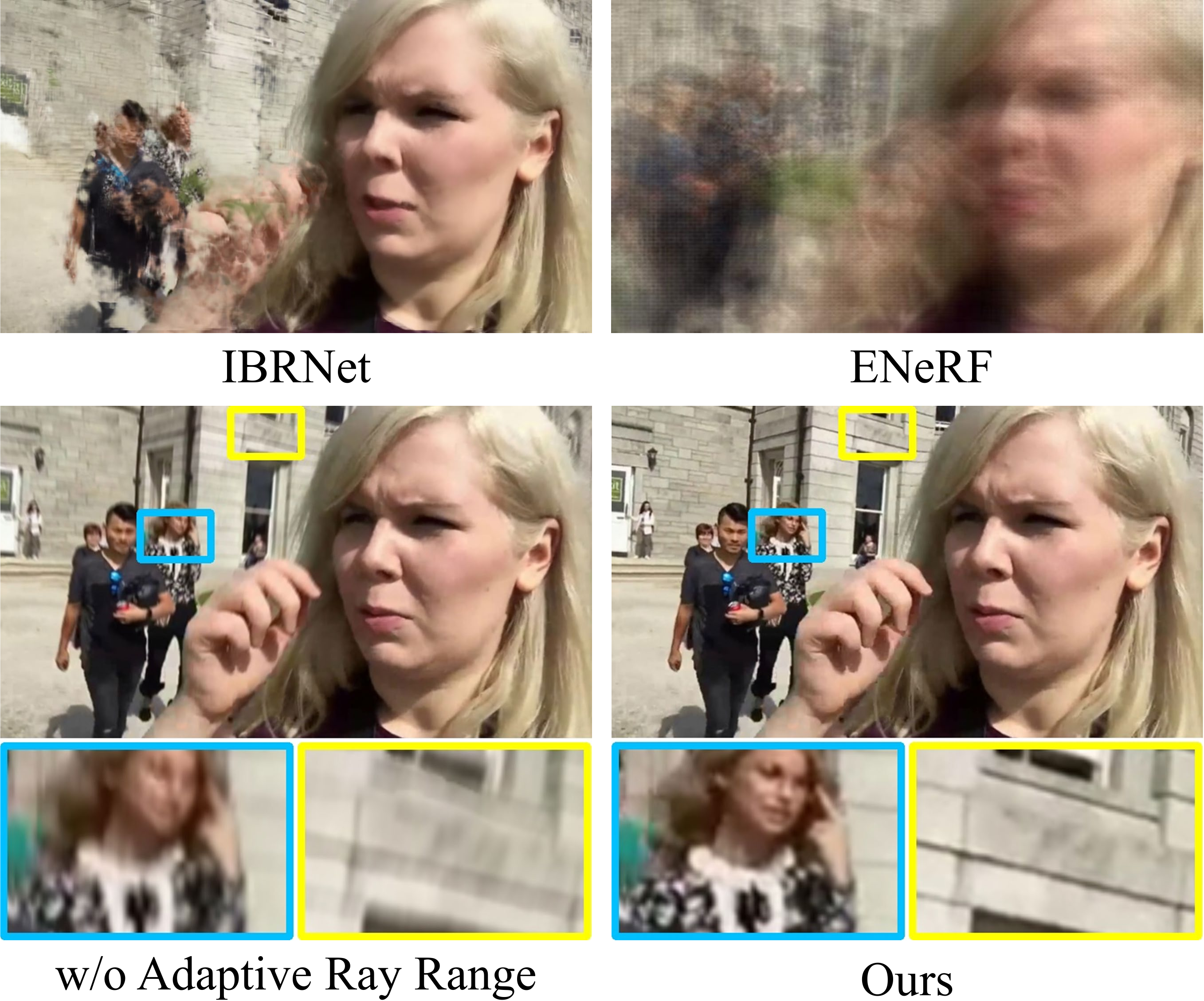}

   \caption{\textbf{Quilitative ablation of different range strategies.}~
   Among the range strategies examined, only our Adaptive Ray Range module can address distortion in image structure.
   } 
   \label{fig:adaptive_ray_range}
\end{figure}

\noindent\textbf{Importance of Adaptive Ray Range.}~
We compare various range strategies to affirm the importance of ARR:~(1)~IBRNet~\cite{wang2021ibrnet} and ENeRF~\cite{lin2022enerf} employ coarse-to-fine range strategy, and~(2)~we adopt even sampling of 128 points following setting of IBRNet as a substitution for ARR. However, as shown in Fig.~\ref{fig:adaptive_ray_range}, none of these strategies achieve favorable results. Without the sampling range defined by ARR, the methods above are forced to aggregate points sampled over a large range, increasing the risk of projecting spatial points onto dynamic regions. Due to the violation of epipolar constraints, dynamic regions introduce incorrect features and colors to the aggregation of descriptors and lead to distortion of the structure. As shown in Row~1,2,3,5 of Table~\ref{tab:quantitative}, ARR proves effective in preserving structure.

\begin{figure}[t]
   \includegraphics[width=1\linewidth]{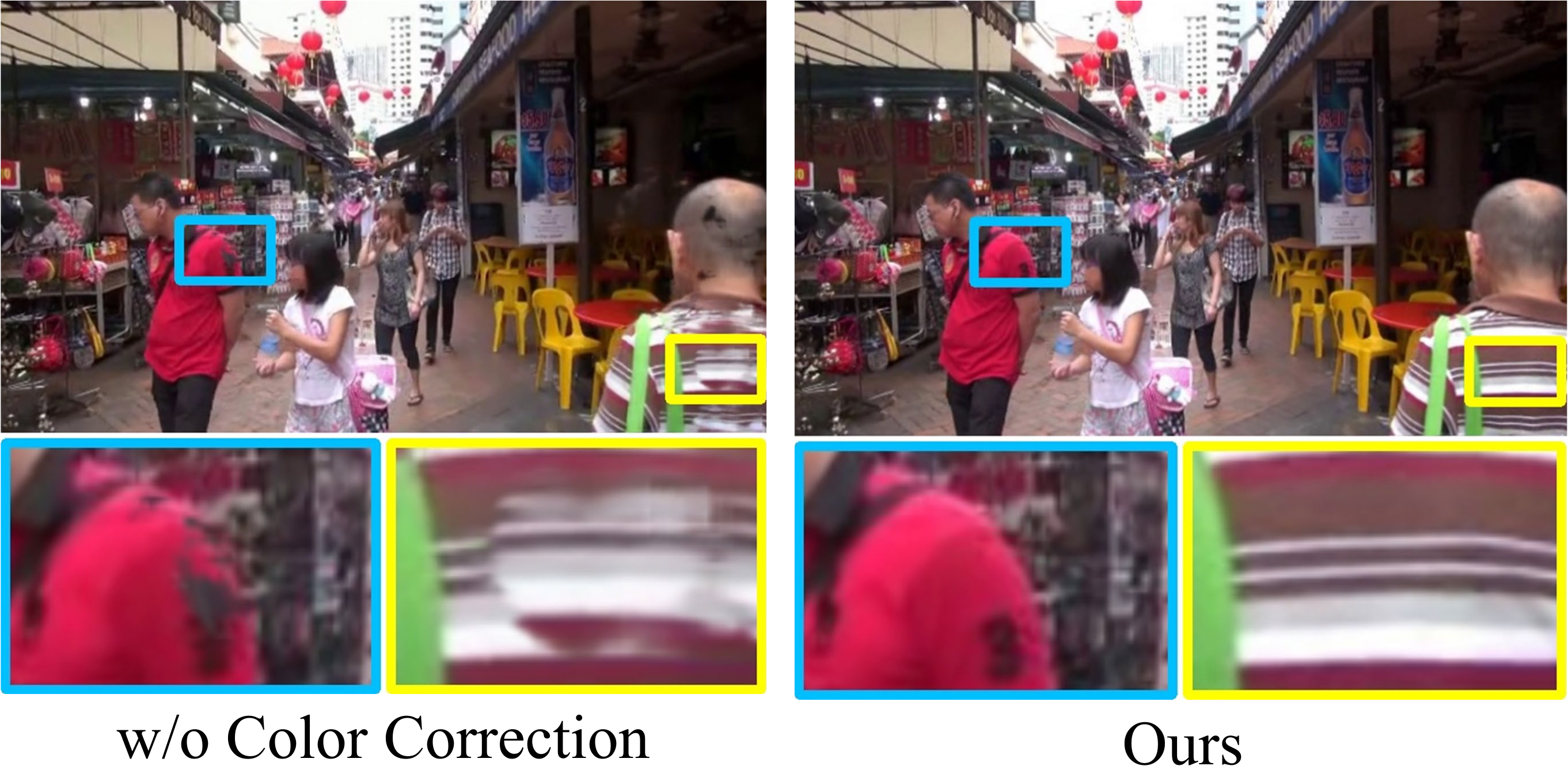}

   \caption{\textbf{Quilitative ablation of Color Correction.}~The Color Correction module refining the projection enhances color accuracy, consequently reducing image artifacts. } 
   \label{fig:color_correction}
\end{figure}

\vspace{5pt}
\noindent\textbf{Importance of Color Correction.}~We conduct a comparison between the results obtained by removing CC and using the full model. The presence of noticeable artifacts in the dynamic region in the left part of Fig.~\ref{fig:color_correction} leads to the decrease in PSNR comparing Row~6 and Row~3 of Table~\ref{tab:quantitative}. This suggests that employing optical flow in CC to refine the projection can improve color accuracy.

\begin{table}\small

\centering

\begin{tabular}{@{}lcc@{}}
\toprule
Method                          & Distortion Value$\uparrow$ & PSNR$\uparrow$  \\ \midrule
ENeRF                           &  -               &      13.45    \\
IBRNet                          &  0.80            & 28.31  \\
Full~(Ours)                      &  \textbf{0.90}           & \textbf{40.01}   \\ \midrule
w/o Stabilized Rendering           &  0.87            & 23.56 \\
w/o Adaptive Ray Range          &  0.81            & 37.83   \\
w/o Color Correction            &  0.86            & 35.81  \\
\bottomrule
\end{tabular}
\caption{\textbf{Quantitative results of ablation study}. We conduct comparative experiments of various range strategies and study the effect of each module. It should be noted that the results of ENeRF are so poor that the Distortion Value is unavailable.}
\label{ablation}
   \vspace{-5pt}
\end{table}

\section{Conclusions}
In this paper, we propose a video stabilization framework termed RStab for integrating multi-frame fusion and $3$D constraints to achieve full-frame generation and structure preservation. The core of RStab lies in \textbf{S}tabilized \textbf{R}endering, a volume rendering module utilizing both colors and features for multi-frame fusion in 3D space. 
To enhance \textbf{S}tabilized \textbf{R}endering module, we design an Adaptive Ray Range module for suppressing inconsistent information and a Color Correction module for refining color aggregation.
By applying the three modules, RStab achieves full-frame generation with structure preservation and outperforms all previous stabilizers in FOV, image quality, and video stability across various datasets.

{
    \small
    \bibliographystyle{ieeenat_fullname}
    \bibliography{main}

\begin{thebibliography}{42}
\providecommand{\natexlab}[1]{#1}
\providecommand{\url}[1]{\texttt{#1}}
\expandafter\ifx\csname urlstyle\endcsname\relax
  \providecommand{\doi}[1]{doi: #1}\else
  \providecommand{\doi}{doi: \begingroup \urlstyle{rm}\Url}\fi

\bibitem[Barron et~al.(2021)Barron, Mildenhall, Tancik, Hedman, Martin-Brualla, and Srinivasan]{barron2021mip}
Jonathan~T Barron, Ben Mildenhall, Matthew Tancik, Peter Hedman, Ricardo Martin-Brualla, and Pratul~P Srinivasan.
\newblock Mip-nerf: A multiscale representation for anti-aliasing neural radiance fields.
\newblock In \emph{Proceedings of IEEE International Conference on Computer Vision (ICCV)}, pages 5855--5864, 2021.

\bibitem[Barron et~al.(2022)Barron, Mildenhall, Verbin, Srinivasan, and Hedman]{barron2022mip360}
Jonathan~T Barron, Ben Mildenhall, Dor Verbin, Pratul~P Srinivasan, and Peter Hedman.
\newblock Mip-nerf 360: Unbounded anti-aliased neural radiance fields.
\newblock In \emph{Proceedings of IEEE Conference on Computer Vision Pattern Recognition (CVPR)}, pages 5470--5479, 2022.

\bibitem[Chen et~al.(2021{\natexlab{a}})Chen, Xu, Zhao, Zhang, Xiang, Yu, and Su]{chen2021mvsnerf}
Anpei Chen, Zexiang Xu, Fuqiang Zhao, Xiaoshuai Zhang, Fanbo Xiang, Jingyi Yu, and Hao Su.
\newblock Mvsnerf: Fast generalizable radiance field reconstruction from multi-view stereo.
\newblock In \emph{Proceedings of IEEE International Conference on Computer Vision (ICCV)}, pages 14124--14133, 2021{\natexlab{a}}.

\bibitem[Chen et~al.(2021{\natexlab{b}})Chen, Tseng, Lee, Chen, and Hung]{chen2021pixstab}
Yu{-}Ta Chen, Kuan{-}Wei Tseng, Yao{-}Chih Lee, Chun{-}Yu Chen, and Yi{-}Ping Hung.
\newblock Pixstabnet: Fast multi-scale deep online video stabilization with pixel-based warping.
\newblock In \emph{Proceedings of IEEE International Conference on Image Processing (ICIP)}, pages 1929--1933, 2021{\natexlab{b}}.

\bibitem[Choi and Kweon(2020)]{choi2020difrint}
Jinsoo Choi and In~So Kweon.
\newblock Deep iterative frame interpolation for full-frame video stabilization.
\newblock \emph{ACM Transactions on Graphics (TOG)}, 39\penalty0 (1):\penalty0 4:1--4:9, 2020.

\bibitem[Goldstein and Fattal(2012)]{goldstein2012epipolar}
Amit Goldstein and Raanan Fattal.
\newblock Video stabilization using epipolar geometry.
\newblock \emph{ACM Transactions on Graphics (TOG)}, 31\penalty0 (5):\penalty0 126:1--126:10, 2012.

\bibitem[Grundmann et~al.(2011)Grundmann, Kwatra, and Essa]{grundmann2011l1}
Matthias Grundmann, Vivek Kwatra, and Irfan~A. Essa.
\newblock Auto-directed video stabilization with robust {L1} optimal camera paths.
\newblock In \emph{Proceedings of IEEE Conference on Computer Vision Pattern Recognition (CVPR)}, pages 225--232, 2011.

\bibitem[Kingma and Ba(2015)]{kingma2015adam}
Diederik~P. Kingma and Jimmy Ba.
\newblock Adam: {A} method for stochastic optimization.
\newblock In \emph{Proceedings of International Conference on Learning Representations (ICLR)}, 2015.

\bibitem[Koh et~al.(2015)Koh, Lee, and Kim]{Yeong2015}
Yeong~Jun Koh, Chulwoo Lee, and Chang{-}Su Kim.
\newblock Video stabilization based on feature trajectory augmentation and selection and robust mesh grid warping.
\newblock \emph{IEEE Transactions on Image Processing (TIP)}, 24\penalty0 (12):\penalty0 5260--5273, 2015.

\bibitem[Lee et~al.(2009)Lee, Chuang, Chen, and Ouhyoung]{lee2009feature}
Ken{-}Yi Lee, Yung{-}Yu Chuang, Bing{-}Yu Chen, and Ming Ouhyoung.
\newblock Video stabilization using robust feature trajectories.
\newblock In \emph{Proceedings of IEEE International Conference on Computer Vision (ICCV)}, pages 1397--1404, 2009.

\bibitem[Lee et~al.(2021)Lee, Tseng, Chen, Chen, Chen, and Hung]{lee2021deep3d}
Yao{-}Chih Lee, Kuan{-}Wei Tseng, Yu{-}Ta Chen, Chien{-}Cheng Chen, Chu{-}Song Chen, and Yi{-}Ping Hung.
\newblock 3d video stabilization with depth estimation by cnn-based optimization.
\newblock In \emph{Proceedings of IEEE Conference on Computer Vision Pattern Recognition (CVPR)}, pages 10621--10630, 2021.

\bibitem[Li et~al.(2023{\natexlab{a}})Li, Song, Chen, Xie, and Zhang]{li2023imu}
Chen Li, Li Song, Shuai Chen, Rong Xie, and Wenjun Zhang.
\newblock Deep online video stabilization using {IMU} sensors.
\newblock \emph{IEEE Transactions on Multimedia (TMM)}, 25:\penalty0 2047--2060, 2023{\natexlab{a}}.

\bibitem[Li et~al.(2021)Li, Niklaus, Snavely, and Wang]{li2021nsff}
Zhengqi Li, Simon Niklaus, Noah Snavely, and Oliver Wang.
\newblock Neural scene flow fields for space-time view synthesis of dynamic scenes.
\newblock In \emph{Proceedings of IEEE Conference on Computer Vision Pattern Recognition (CVPR)}, pages 6498--6508, 2021.

\bibitem[Li et~al.(2023{\natexlab{b}})Li, Wang, Cole, Tucker, and Snavely]{li2023dynibar}
Zhengqi Li, Qianqian Wang, Forrester Cole, Richard Tucker, and Noah Snavely.
\newblock Dynibar: Neural dynamic image-based rendering.
\newblock In \emph{Proceedings of IEEE Conference on Computer Vision Pattern Recognition (CVPR)}, pages 4273--4284, 2023{\natexlab{b}}.

\bibitem[Lin et~al.(2022)Lin, Peng, Xu, Yan, Shuai, Bao, and Zhou]{lin2022enerf}
Haotong Lin, Sida Peng, Zhen Xu, Yunzhi Yan, Qing Shuai, Hujun Bao, and Xiaowei Zhou.
\newblock Efficient neural radiance fields for interactive free-viewpoint video.
\newblock In \emph{ACM SIGGRAPH Asia}, pages 39:1--39:9, 2022.

\bibitem[Lin et~al.(2017)Lin, Jiang, Liu, Cheong, Do, and Lu]{jiang2017deformation}
Kaimo Lin, Nianjuan Jiang, Shuaicheng Liu, Loong{-}Fah Cheong, Minh~N. Do, and Jiangbo Lu.
\newblock Direct photometric alignment by mesh deformation.
\newblock In \emph{Proceedings of IEEE Conference on Computer Vision Pattern Recognition (CVPR)}, pages 2701--2709, 2017.

\bibitem[Liu et~al.(2009)Liu, Gleicher, Jin, and Agarwala]{liu2009content}
Feng Liu, Michael Gleicher, Hailin Jin, and Aseem Agarwala.
\newblock Content-preserving warps for 3d video stabilization.
\newblock \emph{ACM Transactions on Graphics (TOG)}, 28\penalty0 (3):\penalty0 44, 2009.

\bibitem[Liu et~al.(2011)Liu, Gleicher, Wang, Jin, and Agarwala]{liu2011subspace}
Feng Liu, Michael Gleicher, Jue Wang, Hailin Jin, and Aseem Agarwala.
\newblock Subspace video stabilization.
\newblock \emph{ACM Transactions on Graphics (TOG)}, 30\penalty0 (1):\penalty0 4:1--4:10, 2011.

\bibitem[Liu et~al.(2012)Liu, Wang, Yuan, Bu, Tan, and Sun]{liu2012depth}
Shuaicheng Liu, Yinting Wang, Lu Yuan, Jiajun Bu, Ping Tan, and Jian Sun.
\newblock Video stabilization with a depth camera.
\newblock In \emph{Proceedings of IEEE Conference on Computer Vision Pattern Recognition (CVPR)}, pages 89--95, 2012.

\bibitem[Liu et~al.(2013)Liu, Yuan, Tan, and Sun]{liu2013bundle}
Shuaicheng Liu, Lu Yuan, Ping Tan, and Jian Sun.
\newblock Bundled camera paths for video stabilization.
\newblock \emph{ACM Transactions on Graphics (TOG)}, 32\penalty0 (4):\penalty0 78:1--78:10, 2013.

\bibitem[Liu et~al.(2014)Liu, Yuan, Tan, and Sun]{liu2014steadyflow}
Shuaicheng Liu, Lu Yuan, Ping Tan, and Jian Sun.
\newblock Steadyflow: Spatially smooth optical flow for video stabilization.
\newblock In \emph{Proceedings of IEEE Conference on Computer Vision Pattern Recognition (CVPR)}, pages 4209--4216, 2014.

\bibitem[Liu et~al.(2016)Liu, Tan, Yuan, Sun, and Zeng]{liu2016meshflow}
Shuaicheng Liu, Ping Tan, Lu Yuan, Jian Sun, and Bing Zeng.
\newblock Meshflow: Minimum latency online video stabilization.
\newblock In \emph{Proceedings of European Conference on Computer Vision (ECCV)}, pages 800--815, 2016.

\bibitem[Liu et~al.(2017)Liu, Li, Zhu, and Zeng]{liu2017codingflow}
Shuaicheng Liu, Mingyu Li, Shuyuan Zhu, and Bing Zeng.
\newblock Codingflow: Enable video coding for video stabilization.
\newblock \emph{IEEE Transactions on Image Processing (TIP)}, 26\penalty0 (7):\penalty0 3291--3302, 2017.

\bibitem[Liu et~al.(2021)Liu, Lai, Yang, Chuang, and Huang]{liu2021fusta}
Yu{-}Lun Liu, Wei{-}Sheng Lai, Ming{-}Hsuan Yang, Yung{-}Yu Chuang, and Jia{-}Bin Huang.
\newblock Hybrid neural fusion for full-frame video stabilization.
\newblock In \emph{Proceedings of IEEE International Conference on Computer Vision (ICCV)}, pages 2279--2288, 2021.

\bibitem[Martin-Brualla et~al.(2021)Martin-Brualla, Radwan, Sajjadi, Barron, Dosovitskiy, and Duckworth]{martin2021nerfwild}
Ricardo Martin-Brualla, Noha Radwan, Mehdi~SM Sajjadi, Jonathan~T Barron, Alexey Dosovitskiy, and Daniel Duckworth.
\newblock Nerf in the wild: Neural radiance fields for unconstrained photo collections.
\newblock In \emph{Proceedings of IEEE Conference on Computer Vision Pattern Recognition (CVPR)}, pages 7210--7219, 2021.

\bibitem[Meuleman et~al.(2023)Meuleman, Liu, Gao, Huang, Kim, Kim, and Kopf]{meuleman2023progressive}
Andreas Meuleman, Yu{-}Lun Liu, Chen Gao, Jia{-}Bin Huang, Changil Kim, Min~H. Kim, and Johannes Kopf.
\newblock Progressively optimized local radiance fields for robust view synthesis.
\newblock In \emph{Proceedings of IEEE Conference on Computer Vision Pattern Recognition (CVPR)}, pages 16539--16548, 2023.

\bibitem[Mildenhall et~al.(2019)Mildenhall, Srinivasan, Cayon, Kalantari, Ramamoorthi, Ng, and Kar]{mildenhall2019llff}
Ben Mildenhall, Pratul~P. Srinivasan, Rodrigo~Ortiz Cayon, Nima~Khademi Kalantari, Ravi Ramamoorthi, Ren Ng, and Abhishek Kar.
\newblock Local light field fusion: practical view synthesis with prescriptive sampling guidelines.
\newblock \emph{ACM Transactions on Graphics (TOG)}, 38\penalty0 (4):\penalty0 29:1--29:14, 2019.

\bibitem[M{\"u}ller et~al.(2022)M{\"u}ller, Evans, Schied, and Keller]{muller2022instant}
Thomas M{\"u}ller, Alex Evans, Christoph Schied, and Alexander Keller.
\newblock Instant neural graphics primitives with a multiresolution hash encoding.
\newblock \emph{ACM Transactions on Graphics (TOG)}, 41\penalty0 (4):\penalty0 1--15, 2022.

\bibitem[Niklaus and Liu(2020)]{niklaus2020softsplat}
Simon Niklaus and Feng Liu.
\newblock Softmax splatting for video frame interpolation.
\newblock In \emph{Proceedings of IEEE Conference on Computer Vision Pattern Recognition (CVPR)}, pages 5436--5445, 2020.

\bibitem[Shi et~al.(2022)Shi, Shi, Lai, Liang, and Liang]{shi2022fused}
Zhenmei Shi, Fuhao Shi, Wei{-}Sheng Lai, Chia{-}Kai Liang, and Yingyu Liang.
\newblock Deep online fused video stabilization.
\newblock In \emph{Proceedings of Winter Conference on Applications of Computer Vision (WACV)}, pages 865--873. {IEEE}, 2022.

\bibitem[Smith et~al.(2009)Smith, Zhang, Jin, and Agarwala]{smith2009light}
Brandon~M. Smith, Li Zhang, Hailin Jin, and Aseem Agarwala.
\newblock Light field video stabilization.
\newblock In \emph{Proceedings of IEEE International Conference on Computer Vision (ICCV)}, pages 341--348, 2009.

\bibitem[Trevithick and Yang(2021)]{trevithick2021grf}
Alex Trevithick and Bo Yang.
\newblock Grf: Learning a general radiance field for 3d representation and rendering.
\newblock In \emph{Proceedings of IEEE International Conference on Computer Vision (ICCV)}, pages 15182--15192, 2021.

\bibitem[Wang et~al.(2019)Wang, Yang, Lin, Zhang, Shamir, Lu, and Hu]{wang2019deepstab}
Miao Wang, Guo{-}Ye Yang, Jin{-}Kun Lin, Song{-}Hai Zhang, Ariel Shamir, Shao{-}Ping Lu, and Shi{-}Min Hu.
\newblock Deep online video stabilization with multi-grid warping transformation learning.
\newblock \emph{IEEE Transactions on Image Processing (TIP)}, 28\penalty0 (5):\penalty0 2283--2292, 2019.

\bibitem[Wang et~al.(2021)Wang, Wang, Genova, Srinivasan, Zhou, Barron, Martin{-}Brualla, Snavely, and Funkhouser]{wang2021ibrnet}
Qianqian Wang, Zhicheng Wang, Kyle Genova, Pratul~P. Srinivasan, Howard Zhou, Jonathan~T. Barron, Ricardo Martin{-}Brualla, Noah Snavely, and Thomas~A. Funkhouser.
\newblock Ibrnet: Learning multi-view image-based rendering.
\newblock In \emph{Proceedings of IEEE Conference on Computer Vision Pattern Recognition (CVPR)}, pages 4690--4699, 2021.

\bibitem[Xu et~al.(2022{\natexlab{a}})Xu, Xu, Philip, Bi, Shu, Sunkavalli, and Neumann]{xu2022pointner}
Qiangeng Xu, Zexiang Xu, Julien Philip, Sai Bi, Zhixin Shu, Kalyan Sunkavalli, and Ulrich Neumann.
\newblock Point-nerf: Point-based neural radiance fields.
\newblock In \emph{Proceedings of IEEE Conference on Computer Vision Pattern Recognition (CVPR)}, pages 5438--5448, 2022{\natexlab{a}}.

\bibitem[Xu et~al.(2022{\natexlab{b}})Xu, Zhang, Maybank, and Tao]{xu2022dut}
Yufei Xu, Jing Zhang, Stephen~J. Maybank, and Dacheng Tao.
\newblock {DUT:} learning video stabilization by simply watching unstable videos.
\newblock \emph{IEEE Transactions on Image Processing (TIP)}, 31:\penalty0 4306--4320, 2022{\natexlab{b}}.

\bibitem[Yu et~al.(2021)Yu, Ye, Tancik, and Kanazawa]{yu2021pixelnerf}
Alex Yu, Vickie Ye, Matthew Tancik, and Angjoo Kanazawa.
\newblock pixelnerf: Neural radiance fields from one or few images.
\newblock In \emph{Proceedings of IEEE Conference on Computer Vision Pattern Recognition (CVPR)}, pages 4578--4587, 2021.

\bibitem[Yu and Ramamoorthi(2018)]{yu2018selfie}
Jiyang Yu and Ravi Ramamoorthi.
\newblock Selfie video stabilization.
\newblock In \emph{Proceedings of European Conference on Computer Vision (ECCV)}, pages 569--584, 2018.

\bibitem[Yu and Ramamoorthi(2019)]{yu2019robust}
Jiyang Yu and Ravi Ramamoorthi.
\newblock Robust video stabilization by optimization in {CNN} weight space.
\newblock In \emph{Proceedings of IEEE Conference on Computer Vision Pattern Recognition (CVPR)}, pages 3800--3808, 2019.

\bibitem[Yu and Ramamoorthi(2020)]{yu2020stable}
Jiyang Yu and Ravi Ramamoorthi.
\newblock Learning video stabilization using optical flow.
\newblock In \emph{Proceedings of IEEE Conference on Computer Vision Pattern Recognition (CVPR)}, pages 8156--8164, 2020.

\bibitem[Zhao and Ling(2020)]{Zhao2020PWStableNet}
Minda Zhao and Qiang Ling.
\newblock Pwstablenet: Learning pixel-wise warping maps for video stabilization.
\newblock \emph{IEEE Transactions on Image Processing (TIP)}, 29:\penalty0 3582--3595, 2020.

\bibitem[Zhao et~al.(2023)Zhao, Li, Peng, Luo, Ye, Lu, and Cao]{zhao2023fast}
Weiyue Zhao, Xin Li, Zhan Peng, Xianrui Luo, Xinyi Ye, Hao Lu, and Zhiguo Cao.
\newblock Fast full-frame video stabilization with iterative optimization.
\newblock In \emph{Proceedings of IEEE International Conference on Computer Vision (ICCV)}, pages 23534--23544, 2023.

\end{thebibliography}
}

\end{document}